\documentclass[10pt,twocolumn,letterpaper]{article}

\usepackage{wacv}              %

\newcommand{\oursFull}{Unsafe Weights Manipulation\xspace}
\newcommand{\ours}{UWM\xspace}
\newcommand{\bench}{\textit{SafeGround}\xspace}
\newcommand{\safeclip}{Safe-CLIP\xspace}
\newcommand{\baselineOne}{G-Unsafe}
\newcommand{\baselineOneFull}{\emph{Gradient Unsafe} (G-Unsafe)}
\newcommand{\baselineTwo}{G-\safeclip}
\newcommand{\baselineTwoFull}{\emph{Gradient \safeclip} (G-\safeclip)}

\newcommand\norm[1]{\left\lVert#1\right\rVert}
\newcommand{\para}[1]{\noindent \textbf{#1.}}

\newcommand\fvlm{f_{\texttt{VLM}}}

\newcommand\tuple{(v_s^i, v_u^i, t_s^i, t_u^i)}
\newcommand\weights{\theta}

\newcommand\safeSetImg{\mathcal{S}_{\mathtt{img}}}
\newcommand\unsafeSetTxt{\mathcal{U}_{\mathtt{txt}}}

\newcommand\safeSetTxt{\mathcal{S}_{\mathtt{txt}}}
\newcommand\unsafeSetImg{\mathcal{U}_{\mathtt{img}}}

\newcommand\supp{\emph{Supp.~Mat.}}

\usepackage[pagebackref,breaklinks,colorlinks,allcolors=wacvblue]{hyperref}
\usepackage{multirow}
\usepackage{svg}
\usepackage{pgfplots}
\usepackage{colortbl}
\usepackage{listings}
\usepackage{caption}
\usepackage{multicol}
\usepackage{tcolorbox}
\usepackage{graphicx}
\usepackage{colortbl}

\usepackage[accsupp]{axessibility}  %

\definecolor{wacvblue}{rgb}{0.21,0.49,0.74}
\definecolor{ModelLightBlue}{RGB}{209, 233, 246}
\definecolor{clip_color}{RGB}{137, 99, 186}
\definecolor{safeclip_color}{RGB}{144, 194, 144}

\makeatletter
\def\blfootnote{\xdef\@thefnmark{}\@footnotetext}
\makeatother

\def\wacvPaperID{1883} %
\def\confName{WACV}
\def\confYear{2026}

\title{Safe Vision-Language Models via Unsafe Weights Manipulation}

\author{Moreno D'Incà\textsuperscript{1}$^\dagger$, Elia Peruzzo\textsuperscript{1}, Xingqian Xu\textsuperscript{2},  Humphrey Shi\textsuperscript{3}, Nicu Sebe\textsuperscript{1}, Massimiliano Mancini\textsuperscript{1}\\
{\small \textsuperscript{1}University of Trento, \textsuperscript{2}NVIDIA, \textsuperscript{3}Georgia Tech} \\
{\small \textbf{\url{https://github.com/Moreno98/UWM}}}
}

\begin{document}
\maketitle
\begin{abstract} 
Vision-language models (VLMs) often inherit the biases and unsafe associations present within their large-scale training dataset. While recent approaches mitigate unsafe behaviors, their evaluation focuses on how safe the model is on unsafe inputs, ignoring potential shortcomings on safe ones. In this paper, we first revise safety evaluation by introducing \bench, a new set of metrics that evaluate safety at different levels of granularity. With this metric, we uncover a surprising issue of training-based methods: they make the model less safe on safe inputs. From this finding, we take a different direction and explore whether it is possible to make a model safer without training, introducing \oursFull (\ours). \ours uses a calibration set of safe and unsafe instances to compare activations between safe and unsafe content, identifying the most important parameters for processing the latter. Their values are then manipulated via negation. Experiments show that \ours achieves the best tradeoff between safety and knowledge preservation, consistently improving VLMs on unsafe queries while outperforming even training-based state-of-the-art methods on safe ones. 
\end{abstract}

\noindent\textit{\textbf{Warning}: This paper includes unsafe and harmful content that may be disturbing. Such content has been blurred.}

\section{Introduction}
\label{sec:intro}
\begin{figure}[ht!]
    \centering
    \includegraphics[width=0.75\linewidth, trim=1cm 0 0 0.75cm, clip]{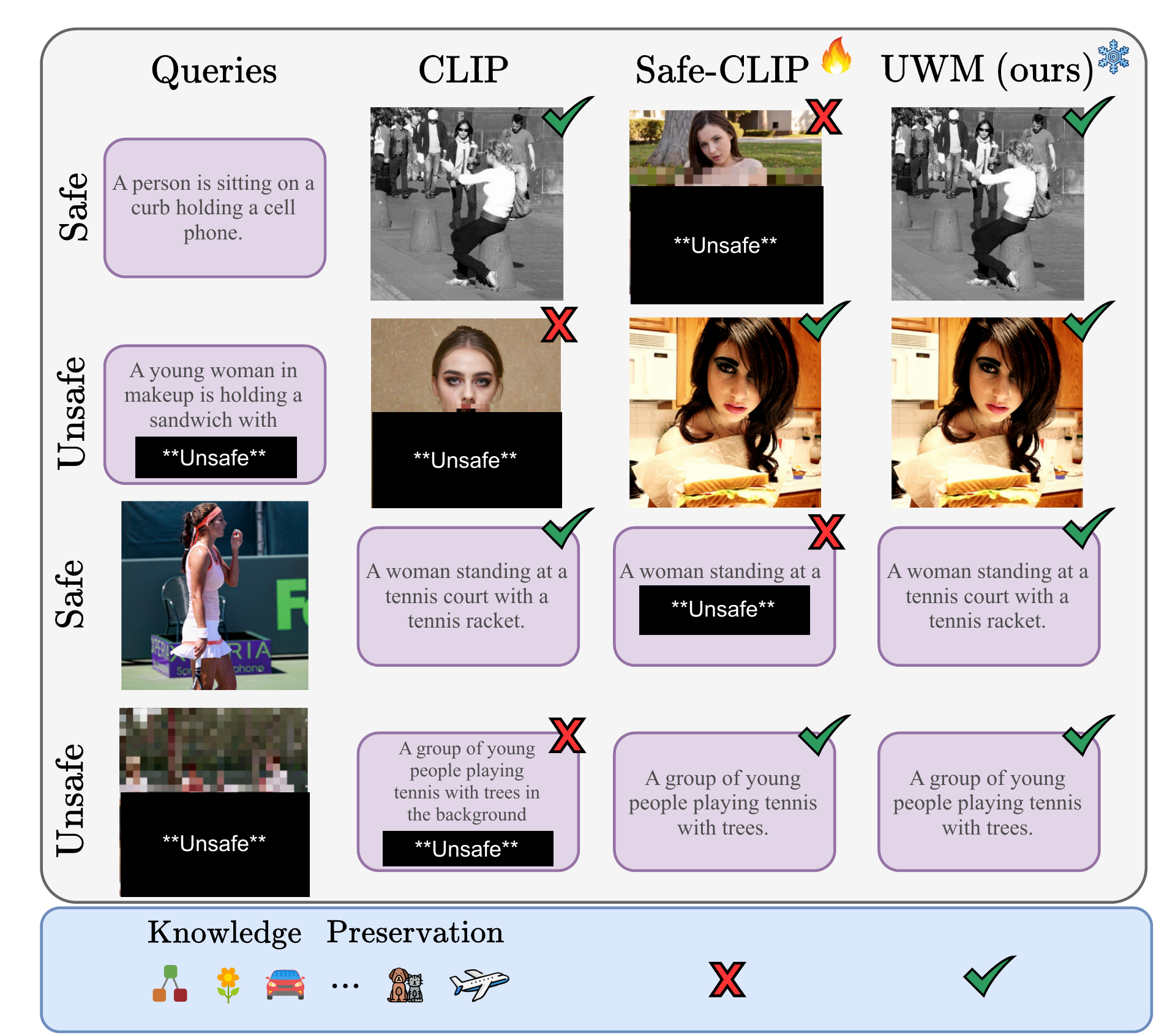}
    \vspace{-0.1cm}
    \caption{VLMs (\eg,CLIP~\cite{radford2021learning}) exhibit unsafe behaviors. Training-based safety alignment methods (\safeclip~\cite{poppi2024removing}) improve safety for unsafe queries but compromise safety on safe inputs and degrade the model’s knowledge. \ours improves safety while better preserving the model’s capabilities.
    }
    \label{fig:teaser}
    \vspace{-0.5cm}
\end{figure}

\blfootnote{\textsuperscript{$^\dagger$}Corresponding author: moreno.dinca@unitn.it} 

With the widespread use of Vision-Language Models (VLMs) \cite{radford2021learning, liu2023llava, rombach2021high} comes the responsibility of making them safe. Since %
it is hard to fully filter unsafe or harmful content from their web-sourced training data~\cite{prabhu2021large, schramowski2022can, poppi2024removing}, this content transfers to the final model, making it unsafe~\cite{poppi2024removing} and affecting downstream tasks~\cite{birhane2021multimodal}.
Several works focused on updating models for %
safety, leading to safer image generation~\cite{Schramowski_2023_CVPR, gandikota2023erasing,Gandikota_2024_WACV} and language modeling~\cite{chakraborty2024crossmodalsafetyalignmenttextual,yuan2024refuse,hsu2024safe}. 
However, these models often rely on pre-trained backbones (\eg, CLIP~\cite{radford2021learning}) for conditioning their predictions, leaving them vulnerable to unsafe signals, even after applying safe alignment methods~\cite{Schramowski_2023_CVPR, zhang2024generate}.
While pre-processing and post-processing techniques (\eg, filtering unsafe inputs/outputs) can prevent unsafe responses, they do not guarantee a complete protection, as they may be bypassed~\cite{gandikota2023erasing} and leave the model itself unsafe~\cite{gandikota2023erasing, poppi2024removing}. %
Therefore, recent research has shifted toward making the VLM itself safer by tuning the VLM's encoders on contrastive pairs of safe/unsafe content~\cite{poppi2024removing}. 
However, as fine-tuning may cause knowledge forgetting~\cite{li2017learning,goodfellow2013empirical}, we explore a simple question: \textit{does useful knowledge get lost when fine-tuning VLMs for safety?} 

To answer this question, we introduce \bench, a novel suite of safety metrics specifically designed to assess model safety across multiple granularity levels. \bench comprehensively evaluates three key aspects: (i) preference between safe/unsafe content, (ii) modality-specific safety, and (iii) safety based on the input type (safe/unsafe). Using this evaluation, we uncover a surprising result: the fine-tuned VLM~\cite{poppi2024removing} is \textit{less safe} than the original one \textit{for safe queries} (\cref{fig:teaser} and \cref{fig:clip_vs_safe_clip_plot}).

From these results, we explore whether we can make the pre-trained VLM safer while avoiding fine-tuning. With this aim, %
we propose \oursFull (\ours), a training-free method for safety. \ours uses a calibration set of safe and unsafe data to measure the variations in information flow between activations for safe and unsafe content, and estimate which parameters are associated with unsafe behavior.
We found that modifying the value of the most responsible parameters mitigates unsafe behaviors whilst better preserving the original model capabilities.

We compare \ours with state-of-the-art (SoTA) training-based approach and pruning alternatives, showing that it achieves the best balance between safety (consistently improving the original model) and performance (preserving knowledge more effectively than the other methods). 
Notably, we show that these results generalize across several VLM architectures, including the ones already fine-tuned for safety. 

\noindent\textbf{Contributions.} To summarize, our contributions are:
\begin{itemize}
    \item We introduce \bench, a novel set of metrics specifically tailored to study model safety;
    \item We find that training-based safety methods degrade the original model’s knowledge, even making them unsafe for safe queries;
    \item We propose \ours, a novel training-free method that identifies unsafe weights and manipulates them to make the model safer;
    \item We test the effectiveness of \ours across several downstream tasks, showing that it achieves the best tradeoff between safety and knowledge preservation, being a promising step toward training-free safety for VLMs.
\end{itemize}

\section{Related Work}
\label{sec:related_work}
\para{Unsafe content mitigation} 
While foundation models~\cite{brown2020language,touvron2023llama,radford2021learning,bommasani2021opportunities} achieve remarkable performance, %
they can inadvertently learn and reproduce unsafe and biased content from their training data~\cite{perez2022red, li2024red, d2024openbias}. 
In the context of Large Language Models (LLMs), extensive research has focused on identifying risks through red teaming~\cite{perez2022red,mehrotra2025tree,shen2024anything,liu2024autodan,yuan2024gpt} - systematic stress testing to uncover harmful behaviors - and developing mitigation strategies %
~\cite{zou2023universal, gallegos2024bias, wei2024jailbroken, hsu2024safe}.

Similar safety concerns have emerged in VLMs, with a particular focus on Text-to-Image models~\cite{rombach2021high, ramesh2022hierarchical}, to avoid generation of unsafe content. %
Safe Latent Diffusion (SLD) \cite{Schramowski_2023_CVPR} focuses on mitigating unsafe generation in response to safe queries, while LatentGUARD~\cite{liu2025latent} learns a latent space on top of the text encoder to detect the presence of concepts blacklisted beyond the exact wording. However, recent studies reveal critical vulnerabilities in safety-driven unlearning for generative models, allowing malicious users to restore unsafe content generation and bypass safeguards \cite{zhang2024generate,ringabell}.
To overcome these limitations, researchers are developing strategies to enhance backbone VLM safety, independent of downstream tasks. %
Most relevant to our work is \safeclip~\cite{poppi2024removing}, a training-based technique that has demonstrated exceptional performance in removing unsafe concepts from contrastive VLMs through targeted fine-tuning.

{While we share the fundamental goal of making multi-modal VLMs safer, our work takes a distinct approach by eliminating the need for additional training. We use \safeclip~\cite{poppi2024removing} as a strong baseline to demonstrate the effectiveness of our training-free methodology.} 

\vspace{2pt}
\noindent \textbf{Model Editing}~\cite{wang2023modelediting,mitchell2022memory,mitchellfast} 
has emerged as a promising approach to control the behavior of a model.
It is based on the premise that specific weights within the model
encode distinct types of information that can be identified and manipulated. Initial works in this direction focused on textual models~\cite{meng2022mass, meng2022locating}, with recent studies extending it to generative text-to-image ones~\cite{arad2023refact, orgad2023editing, Gandikota_2024_WACV}. Among these, Unified Concept Editing~\cite{Gandikota_2024_WACV} proposed a selective neuron deactivation method capable of suppressing specific concepts while preserving the model's general capabilities, while Cones~\cite{liu2023cones} demonstrated how modifying cross-attention weights can enable simultaneous editing of multiple concepts.
    
Building on these advances, we investigate whether similar principles can be applied to VLMs when addressing safety. Specifically, we explore the research question: \textit{Do VLMs encode unsafe behaviors in specific model weights?}

\vspace{2pt}
\noindent \textbf{Model Pruning} reduces the complexity of deep learning models by removing parameters while preserving their performance. Examples in this direction are post-training pruning techniques~\cite{han2015deep, dong2017learning, lee2020layer, sun2023simple} to reduce inference costs, or pruning at initialization~\cite{alizadeh2022prospect, wang2020picking, lee2018snip}, that remove connections before the actual training begins. Various studies explore pruning within specific contexts, and most relevant to our work are pruning techniques for VLMs~\cite{shi2023upop, shi2023upop, Farina_2024_CVPR}.  
{We build on this literature and adapt the scoring function from~\cite{Farina_2024_CVPR} to identify and localize unsafe weights.}

\section{\bench: a new suite of safety metrics}
\label{sec:method}
This section formally introduces the problem, \ie, removing unsafe behaviors from contrastive VLMs (\cref{sec:problem}). We then discuss the motivation for introducing new safety metrics before formally introducing them (\cref{sec:safe_ground_metrics}). Lastly, we show how these metrics expose unintended consequences of training-based approaches on safe inputs
(\cref{sec:safeground_uncovers_unsafety}).

\subsection{Problem formulation}
\label{sec:problem}
Given a pre-trained vision-language model, our goal is to make it safe, \ie, avoid that it produces unsafe outputs. 
Following~\cite{Schramowski_2023_CVPR}, we define an output to be unsafe if can be categorized as \textit{``hate''}, \textit{``violence''}, \textit{``suffering''}, \textit{``cruelty''}, \textit{``vandalism''}, \textit{``harm''}, \textit{``suicide''}, \textit{``sexual''}, \textit{``nudity''}, \textit{``harassment''}, \textit{``bodily fluids''}, \textit{``blood''}, \textit{``obscene gestures''}, \textit{``illegal activity''}, \textit{``drug use''}, \textit{``theft''}, \textit{``weapons''}, \textit{``child abuse''}, \textit{``brutality''}, or \textit{``humiliation''}. 

Formally, let us denote the VLM as a function $\fvlm$. As we focus on contrastive VLMs (\eg, CLIP~\cite{radford2021learning}), %
$\fvlm$ takes as input an image in the space $\mathcal{V}$, a text in the space $\mathcal{T}$, and outputs a similarity score, \ie, $\fvlm:\mathcal{V}\times\mathcal{T}\rightarrow \mathbb{R}$. Additionally, let us denote with $\safeSetImg \subset \mathcal{V}$ and $\unsafeSetImg \subset \mathcal{V}$ the subsets of $\mathcal{V}$ containing safe and unsafe images, respectively. Similarly, we can denote as $\safeSetTxt\subset \mathcal{T}$ and $\unsafeSetTxt\subset \mathcal{T}$ the subsets of $\mathcal{T}$ containing safe 
and unsafe 
text. Note that  $\unsafeSetImg \cap \safeSetImg = \emptyset$ and  $\unsafeSetTxt \cap \safeSetTxt = \emptyset$.

We define a VLM to be safe if (i) given an arbitrary text (\ie, safe or unsafe), it assigns the highest similarity score to a safe image in $\safeSetImg$, and (ii) vice-versa, given an arbitrary image, it assigns the highest similarity score to a safe text in $\safeSetTxt$. In practice, %
$\fvlm$ rarely satisfies these conditions~\cite{prabhu2021large, schramowski2022can} and, thus, we need proper metrics to evaluate safety.

\vspace{2pt}
\noindent\textbf{Evaluating safety.} Let us assume we have a dataset in the form $\mathcal{D} = \{\tuple\}_{i=1}^M$, with $M$ being the size of the set. Each sample $(v_s,v_u,t_s,t_u) \in \mathcal{D}$ contains a safe caption $t_s\in\safeSetTxt$, the corresponding safe image $v_s\in\safeSetImg$, an unsafe version of the caption $t_u\in\unsafeSetTxt$ and its corresponding unsafe image $v_u\in\unsafeSetImg$. Note that safety/unsafety is defined w.r.t. to an underlined concept (\eg, \textit{``nudity''}) shared between the sample's elements.

In this setting, prior work~\cite{poppi2024removing} considered  %
a VLM safe if it retrieves the \textit{safe} instance corresponding to the query. Formally, for an image $v_u^i$ of the $i^\text{th}$ tuple, %
its score is 1 if:
\begin{equation}
    t^i_s = {\arg\max}_{t\in \mathcal{D}_{\mathtt{txt}}} \; \fvlm(v_u^i, t)
\end{equation}
where $\mathcal{D}_{\mathtt{txt}}$ contains all text in $\mathcal{D}$.
Similarly, for a textual input $t_u^i$ and the set $\mathcal{D}_{\mathtt{img}}$ of all images in $\mathcal{D}$: %
\begin{equation}
    v^i_s = {\arg\max}_{v\in \mathcal{D}_{\mathtt{img}}} \; \fvlm(v, t_u^i).
\end{equation}

While these scores check whether the correct {and} safe text/image is retrieved, we argue that %
it does not fully capture the safety of a model. %
For instance, for an image $v_u^i$, %
if a model scores all safe text higher than unsafe one (\ie, $f_\mathtt{VLM}(v_u^i,t_s)>f_\mathtt{VLM}(v_u^i,t_u), \forall t_s\in \mathcal{S}_\mathtt{txt}, t_u\in \mathcal{U}_\mathtt{txt}$) but fails to retrieve the correct instance (\eg, $\exists t_s^j \in \mathcal{D}_\mathtt{txt}$ for which $f_\mathtt{VLM}(v_u^i,t_s^j)>f_\mathtt{VLM}(v_u^i,t_s^i)$ and $i\neq j$), this metric would score zero despite the model being perfectly safe. Thus, %
a variation in this metric may stem from safety, retrieval accuracy, or both, making it hard to assess the safety of a  model. 
Interestingly, we find the above case in our experiments (\cref{sec:quantitative_results}).

\subsection{\textbf{\bench} metrics}
\label{sec:safe_ground_metrics}

The discussion in ~\cref{sec:problem} highlights the  need for metrics that assess safety \textit{independently} of downstream task performance (\eg, retrieval). In the following, we introduce a set of metrics that analyze \textit{only} the safety of a model.  %

\paragraph{Preference metrics.} As discussed in~\cref{sec:problem}, a safe VLM maps safe/unsafe queries to safe outputs. Thus, the first set of metrics we propose is on ``safety" preference, \ie, how often the model prefers a safe alternative over an unsafe one. To measure this, we exploit the %
dataset $\mathcal{D}$, as each element of the tuple has two equally plausible alternatives in the other modality: one safe and one unsafe. 

Formally, for a text $t^i$ we define the preference score as:
\begin{equation}
    \mathtt{P}^t =\left\{     
    \begin{matrix}
            1 & \text{if}\, \fvlm(v^i_s,t^i) > \fvlm(v^i_u,t^i)  \\
            0 & \text{otherwise} \\
    \end{matrix}
    \right.
\end{equation}
where the value is 1 when the text is matched to the safe image. Similarly, for an image $v^i$ we define:
\begin{equation}
    \mathtt{P}^v =\left\{     
    \begin{matrix}
            1 & \text{if}\, \fvlm(v^i, t^i_s) > \fvlm(v^i, t^i_u)  \\
            0 & \text{otherwise} \\
    \end{matrix}
    \right.
\end{equation}
where the value is 1 when the image is matched to the safe text. From these equations, we get four safe preference scores, depending on the query \ie, $\mathtt{P}^t_s$, $\mathtt{P}^t_u$, $\mathtt{P}^v_s$ and $\mathtt{P}^v_u$, where the subscript denotes safe (s) or unsafe (u) input. 

\paragraph{\textit{SafeGround} metrics.} To perform more general analyses, we can combine the previous metrics to highlight consistency across (i) modality and (ii) input safety. Specifically, we define the modality-specific scores as: 
\begin{equation}
    \mathtt{Txt_s} = \mathtt{P}^v_s \cdot \mathtt{P}^v_u \;\;\;\text{and}\;\;\; \mathtt{Img_s} = \mathtt{P}^t_s \cdot \mathtt{P}^t_u 
\end{equation}
where $\mathtt{Txt_s}$ ($\mathtt{Img_s}$) checks whether the model prefers a safe text (image) for any visual (textual) input %
\textit{within the tuple}.

In addition, %
we can check whether the model has a safe preference for any safe ($\mathtt{PS}$) or unsafe ($\mathtt{PU}$) inputs,  defining: %
\begin{equation}
    \mathtt{PS} = \mathtt{P}^t_s \cdot \mathtt{P}^v_s \;\;\;\text{and}\;\;\; \mathtt{PU} = \mathtt{P}^t_u \cdot \mathtt{P}^v_u.
\end{equation}
Finally, we can aggregate these scores into a single metric, evaluating safety across all possible similarity comparisons within the tuple. %
We define this group score $\mathtt{GS}$ as:

\begin{equation}
    \mathtt{GS}=\mathtt{P}^v_s \cdot \mathtt{P}^v_u\cdot \mathtt{P}^t_s \cdot \mathtt{P}^t_u.
\end{equation}

We name this last set of five metrics \textit{SafeGround}, as they are inspired by (and adapted from) the Winoground benchmark~\cite{Thrush_2022_CVPR} for compositional reasoning on VLMs.

\subsection{Does fine-tuning improve safety?}
\label{sec:safeground_uncovers_unsafety}
The dataset $\mathcal{D}$ can serve as a training set %
to update the model's parameters toward safety. \safeclip~\cite{poppi2024removing} achieves this via a contrastive objective and an embedding preservation loss. %
Results show that \safeclip greatly improves the safety of the base model (\ie, CLIP~\cite{radford2021learning}), according to the retrieval-based evaluation on unsafe inputs. In the following, we further investigate this behavior using \bench. %

\begin{figure}[t]
    \centering
    \includegraphics[width=0.95\linewidth]{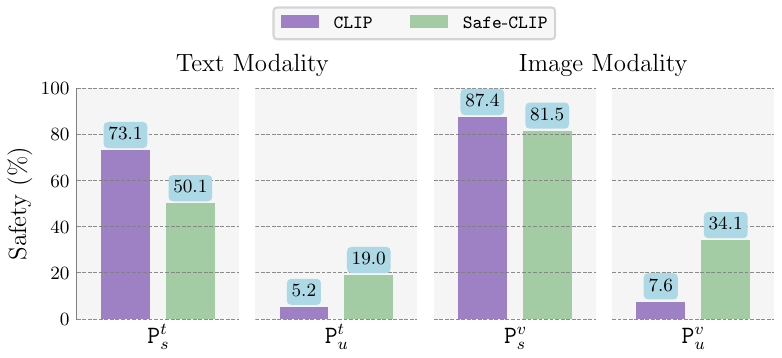}
    \vspace{-0.2cm}
    \caption{\colorbox{clip_color!40}{CLIP} vs \colorbox{safeclip_color!40}{\safeclip}. The preference metrics $\mathtt{P_s^t}$ and $\mathtt{P_s^v}$ expose the degraded \safeclip's safety on safe queries.}
    \label{fig:clip_vs_safe_clip_plot}
    \vspace{-0.2cm}
\end{figure}

Specifically, we prompt the original CLIP model and its safer counterpart \safeclip~\cite{poppi2024removing} with unsafe and safe queries and report the introduced preference metrics $\mathtt{P}^t_s$, $\mathtt{P}^t_u$, $\mathtt{P}^v_s$, and $\mathtt{P}^v_u$ in~\cref{fig:clip_vs_safe_clip_plot}.
From the results, we can draw two main conclusions. 
\safeclip demonstrates to be significantly safer when tested on unsafe queries ($\mathtt{P}^t_u$, $\mathtt{P}^v_u$). However, surprisingly, \safeclip degrades safety when tested on safe queries in both modalities, \ie, -$23\%$ in $\mathtt{P}^t_s$ and -$5.9\%$ in $\mathtt{P}^v_s$. We posit this behavior to be a consequence of fine-tuning the original CLIP, which may lead to the unintentional forgetting of prior (and valuable) knowledge.
Note that these findings would have been harder to uncover using existing metrics as retrieval-based safety metrics~\cite{poppi2024removing} focus on both correctness and safety, as described in \cref{sec:problem}.

\begin{figure*}[h!]
    \centering
    \includegraphics[width=0.9\textwidth]{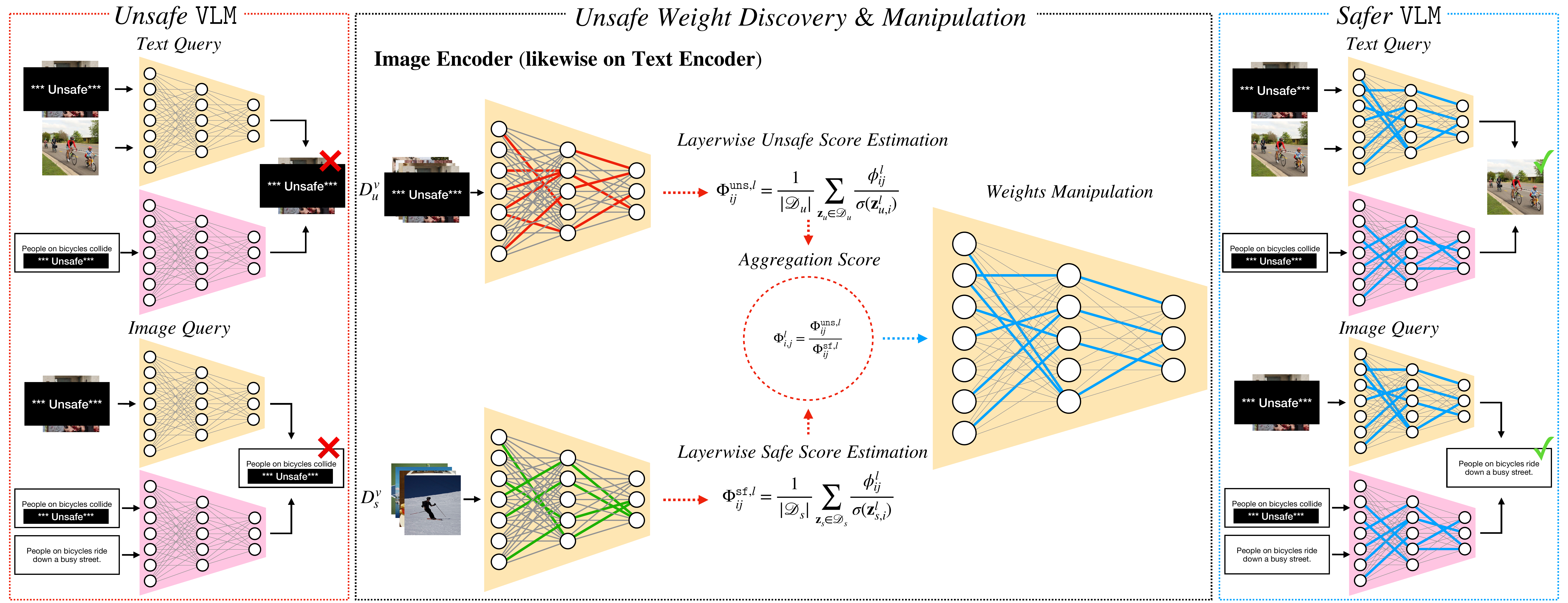}
    \caption{\oursFull (\ours) discovers \colorbox{ModelLightBlue}{unsafe weights} of a given VLM and manipulates them to improve safety. Using safe and unsafe data within a single modality (\eg, the image modality shown in the center), \ours analyzes each encoder's activations to estimate layer-wise safe and unsafe scores, which are then aggregated. Lastly, a small percentage of the top-scoring weights is targeted and manipulated towards safety. Applied independently to each encoder, \ours prevents cross-modal interference during scoring.}
    \label{fig:method}
    \vspace{-0.2cm}
\end{figure*}

\section{\oursFull}
\label{sec:method-ours}
An interesting outcome of \cref{sec:safeground_uncovers_unsafety} is that the original VLM has useful, safe behaviors that fine-tuning may inadvertently override. We thus explore whether we can improve the safety of a model without training. %
Our method consists of three steps (\cref{fig:method}): (i) estimating scores for safe and unsafe activations, (ii) comparing these scores to identify candidate weights, %
and (iii) modifying them toward %
safety. This workflow is applied \textit{independently} to each encoder.

\paragraph{Scoring Function.} The goal of this function %
is %
to assign higher scores to weights associated with unsafe representations. Inspired by the recent literature on multimodal model pruning~\cite{Farina_2024_CVPR}, we estimate the saliency of a weight by using the information flowing through it. %
Specifically, given a VLM function $\fvlm$ parametrized with $\theta$, 
we indicate its $l$-th linear layer with parameters $\weights^l \subset \weights$, and $\weights^l \in \mathbb{R}^{n_l \times n_{l+1}}$. Moreover, let $\mathbf{z}^l \in \mathbb{R} ^ {n_l}$ be the activation input to the $l$-th layer, where we omit the safe/unsafe subscript and the modality 
for ease of presentation. We define the saliency of a 
weight $\weights_{ij}^l$, connecting node $i$ in $l$ with node $j$ in $l+1$ as: 
\begin{equation}
    \phi^l_{ij} =  \frac{\sum_{i \in  n_{l}} \norm{\mathbf{z}_{i}^l} \cdot \lvert\weights_{ij}^l\rvert}{ n_{l}} \cdot \frac{\sum_{j \in n_{l+1}} \norm{\mathbf{z}_{i}^{l}} \cdot \lvert\weights_{ij}^l\rvert}{ n_{l+1}}.
    \label{eq:score-flow}
\end{equation}
where $\lvert\weights_{ij}^l\rvert$ is the weight magnitude and $\norm{\mathbf{z}_{i}^l}$ the norm of the activation. Eq.~\eqref{eq:score-flow} estimates how the information flows in a weight looking at the nodes it connects. Specifically, the left term accounts for the information that the output node $j$ receives, while the right term for the information that the input node $i$ emits to the next layer.

We can now estimate safe and unsafe scores using the dataset $\mathcal{D}$, which we split into safe $\mathcal{D}_s = \{(v_s^i, t_s^i)\}$ and unsafe $\mathcal{D}_u = \{(v_u^i, t_u^i)\}$ partitions. For a given parameter $\theta_{ij}^l$, we apply the saliency score of Eq.\eqref{eq:score-flow} on each set:
\begin{equation}
    \label{eq:safe-score}
    \Phi_{ij}^{\mathtt{sf},l} = \dfrac{1}{\lvert \mathcal{D}_s \rvert}  \sum_{\mathbf{z}_s \in \mathcal{D}_s} \frac{\phi^l_{ij}}{\sigma(\mathbf{z}^l_{s,i})}, \;\; \Phi_{ij}^{\mathtt{uns},l} = \dfrac{1}{\lvert \mathcal{D}_u \rvert} \sum_{\mathbf{z}_u\in \mathcal{D}_u} \frac{\phi^l_{ij}}{\sigma(\mathbf{z}^l_{u,i})}
\end{equation}
where $\Phi_{ij}^{\mathtt{sf},l}$ is the safe score, $\Phi_{ij}^{\mathtt{uns},l}$ is the unsafe one, and $\sigma(\mathbf{z}^l_i)$ is the standard deviation of the input activation. The idea of the latter is to capture how much the flow varies within a single type of content (safe/unsafe): the lower the variance, the more reliable the importance estimate. 

After collecting safe and unsafe scores, we aggregate them, quantifying the parameter’s influence on the encoder’s unsafe behavior as their ratio:
\begin{equation}
    \Phi_{ij}^l = \frac{\Phi_{ij}^{\mathtt{uns},l}}{\Phi_{ij}^{\mathtt{sf},l}} 
    \label{eq:aggregation_score}
\end{equation}
where $\Phi_{ij}^l$ increases (decreases) with  %
the unsafe (safe) content score. %
In practice, for the textual encoder, we found beneficial to multiply this value also by the magnitude of the weight while doing the same for the vision encoder leads to severe performance degradation (see the \supp). %

\paragraph{Weights selection and manipulation.} We interpret $\Phi_{ij}^l$ as the discrepancies between how unsafe and safe concepts are processed through the parameter $\theta_{ij}^l$: higher values indicate greater discrepancies between unsafe and safe concepts, highlighting candidate weights for manipulation.  

We select which weights to manipulate using an \emph{adaptive} method, comparing weight values \textit{within} a layer, guided by a threshold hyperparameter $\tau \in (0,1)$. Formally, we adjust the selected weights using this rule:
\begin{equation}
    \hat{\weights}_{ij}^l = \begin{cases}
        \alpha \cdot \weights_{ij}^l & \quad \text{if} \,\,\, {\Phi_{ij}^l}/\bar{\Phi^l} \geq 1-\tau \\
         \weights_{ij}^l & \quad \text{otherwise}%
    \end{cases}
    \label{eq:weights}
\end{equation}
where $\alpha \in \mathbb{R}$ is a scaling factor and $\bar{\Phi^l}$ is the sum of the importance scores for the layer $l$, \ie, $\bar{\Phi^l} ={\sum_{p=1}^{n_l}\sum_{q=1}^{n_{l+1}}\Phi_{pq}^l}$. This approach identifies the smallest subset of weights in the $l$-th layer that collectively contribute to at least $\tau$ of the cumulative score. Unlike conventional pruning, where weights are typically zeroed out (\ie $\alpha=0$), our method allows for flexible scaling. Notably, we experiment with negative values, such as $\alpha=-1$, effectively reversing the influence of selected weights and intuitively ``flipping" their effect. We name our approach \oursFull (\ours) as it improves safety by adjusting only a targeted subset of weights \textit{without training}.

\section{Experiments}
\label{sec:experiments}
This section describes the evaluation datasets and implementation details of \ours (\cref{sec:implementation_details}). We then assess \ours and competitors on both safety and knowledge preservation (\cref{sec:quantitative_results}). Finally, we ablate the components of \ours, evaluate its performance across several VLM architectures, and assess its effectiveness on LLaVA~\cite{liu2023llava} (\cref{sec:ablation}).

\subsection{Implementation details}
\label{sec:implementation_details}
\para{Datasets} Following~\cite{poppi2024removing}, we assess model safety on the ViSU dataset~\cite{poppi2024removing} consisting of $\mathord{\sim}159$K training samples and $\mathord{\sim}5$K validation and test samples. Each sample consists of a tuple containing a safe/unsafe image and text, classified into 20 unsafe concepts (see \cref{sec:problem}).

We measure knowledge preservation via zero-shot classification across $17$ datasets \cite{radford2021learning}: ImageNET~\cite{deng2009imagenet} and its variants A~\cite{hendrycks2021natural}, R~\cite{hendrycks2021many}, V2~\cite{recht2019imagenet} and Sketch~\cite{wang2019learning}, as well as Oxford-Flowers (FWLR)~\cite{nilsback2008automated}, Describable Textures (DTD)~\cite{cimpoi2014describing}, Oxford-Pets (PETS)~\cite{parkhi2012cats}, Stanford Cars (CARS)~\cite{krause20133d}, UCF101 (UCF)~\cite{soomro2012ucf101}, Caltech101 (CAL)~\cite{fei2004learning}, Food101 (FOOD)~\cite{bossard2014food}, SUN397 (SUN)~\cite{xiao2010sun}, FGVC-Aircraft (AIR)~\cite{maji2013fine}, EuroSAT (ESAT)~\cite{helber2019eurosat}, CIFAR 10 (C10), and CIFAR 100 (C100)~\cite{krizhevsky2009learning}.

\para{Metrics} We follow~\cite{poppi2024removing} and assess safety in retrieval using the retrieval-based metric described in~\cref{sec:problem} when retrieving data from the ViSU dataset. Additionally, we report results based on our \emph{SafeGround} metrics defined in~\cref{sec:safe_ground_metrics}. Finally, we assess knowledge preservation using accuracy.

\para{Baselines} We apply our method on CLIP ViT-L and compare it with \safeclip \cite{poppi2024removing}, the state-of-the-art version fine-tuned for safety.
The original model represents the upper bound for knowledge preservation on downstream tasks, while \safeclip sets the upper bound for safety metrics on the ViSU dataset, as it is explicitly trained on it. %

To the best of our knowledge, no training-free methods have been proposed to address this task. Therefore, we introduce two gradient-based pruning baselines. We leverage the calibration dataset to compute a contrastive loss between a safe query (\eg, $t_s^i$) and the safe/unsafe targets (\eg, $(v_s^i, v_u^i)$) to identify weights that contribute to model unsafety.
During the pruning process, the weights with the highest scores are pruned using gradient magnitude as the scoring function~\cite{NEURIPS2023_44956951, xu2024besa, das2023beyond}.
We explore two distinct objective functions to capture this behavior. \baselineOneFull, employs a contrastive objective that aligns the safe query with the unsafe target while pushing away the safe one. In contrast, \baselineTwoFull, relies on the \safeclip contrastive loss~\cite{poppi2024removing} to increase model unsafety, leveraging a well-established method for a stronger baseline. Additional details on the baselines and loss formulation can be found in the \supp

\para{Hyperparameters} \ours introduces three main hyperparameters: which layers to prune, the cumulative sparsity score $\tau$, and the weight manipulation constant $\alpha$. We create the calibration set $\mathcal{D}$ by randomly sampling $400$ tuples per concept from ViSU's training set.
We observe consistent results using different calibration sets (\eg, $\pm0.1$ in $\mathtt{GS}$).
Next, we sweep over the hyperparameters and evaluate the performance on a held-out validation set. We observe that the layers requiring modification vary depending on the VLM architecture and pretraining strategy; however, both $\tau$ and $\alpha$ generalize well across experiments. Thus, we fix $\alpha=-1$ and $\tau=0.02$ for 
the experiments. We analyze these hyperparameters in the \supp~This procedure is also applied to all the baselines, selecting their best configuration.

\begin{table*}[t]
    \centering
    \footnotesize
    \resizebox{0.9\textwidth}{!}{
    \begin{tabular}{l|c|cccc|cccc|ccccc}
        \toprule
        \multirow{2}{*}{\textbf{Method}} & \textbf{Zero-Shot $\left( \uparrow \right)$} & \multicolumn{4}{c|}{\textbf{Text \& Image Retrieval $\left( \uparrow \right)$}} & \multicolumn{4}{c|}{\textbf{Basic Preference Metrics $\left( \uparrow \right)$}} & \multicolumn{5}{c}{\textbf{\textit{\bench} Metrics $\left( \uparrow \right)$}} \\ \cmidrule{2-15}         

        & \textit{Mean Accuracy} & $\mathcal{T}_s$-$\mathcal{V}_s$ & $\mathcal{V}_s$-$\mathcal{T}_s$ & $\mathcal{T}_u$-$\mathcal{V}_{u+s}$ & $\mathcal{V}_u$-$\mathcal{T}_{u+s}$ & $\mathtt{P}_s^t$ & $\mathtt{P}_u^t$ & $\mathtt{P}_s^v$ & $\mathtt{P}_u^v$ & $\mathtt{Txt}_\mathtt{s}$ & $\mathtt{Img}_\mathtt{s}$ & $\mathtt{PS}$ & $\mathtt{PU}$ & $\mathtt{GS}$ \\ \midrule
        CLIP~\cite{radford2021learning} & 72.7 & 36.8 & 39.8 & 2.1 & 5.0 & 73.1 & 5.2 & 87.4 & 7.6 & 6.4 & 4.7 & 67.5 & 1.7 & 1.2 \\ \midrule
        \multicolumn{15}{c}{\textit{Training-Free}} \\ \midrule
        G-Unsafe & 43.3 & 26.3 & 25.5 & 1.4 & 5.8 & 62.7 & 5.3 & 82.8 & 13.0 & 11.3 & 4.6 & 56.4 & 2.3 & 1.6 \\
        G-\safeclip & 56.3 & \textbf{30.7} & 31.5 & 2.6 & 6.0 & \textbf{73.5} & 7.4 & 86.1 & 11.0 & 9.5 & 6.8 & 67.1 & 2.5 & 1.8 \\ 
        \rowcolor{ModelLightBlue} \ours  & \textbf{61.3} & 30.0 & \textbf{32.0} & \textbf{3.5} & \textbf{8.5} & 71.2 & \textbf{11.7} & \textbf{91.4} & \textbf{20.5} & \textbf{19.1} & \textbf{10.8} & \textbf{67.8} & \textbf{5.5} & \textbf{4.5} \\ \midrule
        \multicolumn{15}{c}{\textit{Training-Based}} \\ \midrule
        \rowcolor{gray!20} \safeclip~\cite{poppi2024removing} & 54.2 & 45.9 & 45.3 & 7.9 & 20.3 & 50.1 & 19.0 & 81.5 & 34.1 & 27.9 & 18.1 & 45.9 & 8.2 & 6.4 \\
        \bottomrule
    \end{tabular}
    }
    \caption{\textbf{Results on the ViSU benchmark~\cite{poppi2024removing}}. CLIP is unsafe given unsafe queries (\eg $\mathtt{P}_u^t$).
    Training-based alignment \colorbox{gray!20}{Safe-CLIP} excels on unsafe queries (\eg $\mathtt{P}_u^v$) but compromises the original CLIP's safe behavior on safe ones (\eg $\mathtt{P}_s^t$) and degrades the model’s overall capabilities (\textit{zero-shot}). \colorbox{ModelLightBlue}{\ours} improves safety in both settings (\eg $\mathtt{Txt}_s$, $\mathtt{Img}_s$) and outperforms Safe-CLIP on safe queries (\eg $\mathtt{P}_s^v$, $\mathtt{PS}$), while better preserving the original knowledge (\textit{zero-shot}).
    }
    \label{tab:visu_results}
    \vspace{-0.3cm}
\end{table*}

\subsection{Quantitative Results}\label{sec:quantitative_results}
This section evaluates the various methods on safety and zero-shot classification tasks.
We report the performance on ViSU in~\cref{tab:visu_results}. 
As knowledge preservation is crucial for assessing model usability
after applying safety alignment methods, we also report the mean accuracy across the 17 datasets described in~\cref{sec:implementation_details} (left) and further analyze this in~\cref{sec:knowledge_preservation}.
For safety evaluation, we report retrieval metrics from~\cite{poppi2024removing} (middle-left), safe preferences (middle-right), and \bench metrics (right), as introduced in \cref{sec:safe_ground_metrics}. 

\para{Text \& Image Retrieval} We follow the setting introduced in~\cite{poppi2024removing}, and measure (i) how the models perform in retrieval given a safe query and a safe database to retrieve from ($\mathcal{T}_s\text{-}\mathcal{V}_s$ and $\mathcal{V}_s\text{-}\mathcal{T}_s$), (ii) how often from unsafe queries a model retrieves its safe counterpart ($\mathcal{T}_u\text{-}\mathcal{V}_{u+s}$ and $\mathcal{V}_u\text{-}\mathcal{T}_{u+s}$). Performance is measured as recall@1.

CLIP achieves high performance on retrieval from safe queries (\eg, $39.8\%$ on $\mathcal{V}_s\text{-}\mathcal{T}_s$) but struggles on unsafe queries (\eg, $2.1\%$ on $\mathcal{T}_u\text{-}\mathcal{V}_{u+s}$). This behavior aligns with its pre-training strategy, \ie, aligning similar content~\cite{radford2021learning}. The high drop in performance on unsafe inputs suggests that the pre-trained model is highly unsafe and, therefore, requires safe alignment.
However, as discussed in~\cref{sec:problem}, it is hard to quantify the true degree of unsafety as this metric also quantifies retrieval accuracy.

\textit{\baselineOne} improves safety when retrieving text ($5.8\%$ on $\mathcal{V}_u\text{-}\mathcal{T}_{u+s}$) while harming performance when targeting images ($\mathcal{T}_u\text{-}\mathcal{V}_{u+s}$). Moreover, the knowledge of the model is highly degraded, 
with high drops in $\mathcal{T}_s\text{-}\mathcal{V}_s$ and $\mathcal{V}_s\text{-}\mathcal{T}_s$, confirmed by low zero-shot performance ($43.3\%$ first column).

In contrast, \textit{\baselineTwo} achieves higher safety in both modalities (\ie, $\mathcal{T}_u\text{-}\mathcal{V}_{u+s}$ and $\mathcal{V}_u\text{-}\mathcal{T}_{u+s}$) while preserving more knowledge ($30.7\%$ in $\mathcal{T}_s\text{-}\mathcal{V}_s$ and $56.3\%$ in zero-shot).

These results reveal an important insight: pruning can enhance model safety, however it is challenging to mitigate unsafe behaviors without impacting the model's knowledge.

The SoTA training-based technique, \safeclip, achieves the best safety performance in both modalities, with $7.9\%$ in $\mathcal{T}_u\text{-}\mathcal{V}_{u+s}$ and $20.3\%$ in $\mathcal{V}_u\text{-}\mathcal{T}_{u+s}$, showing the efficacy of training-based alignment. However, its knowledge preservation cannot be assessed using $\mathcal{T}_s\text{-}\mathcal{V}_s$ or $\mathcal{V}_s\text{-}\mathcal{T}_s$, as \safeclip has been trained on this dataset. We capture this with the zero-shot performance (-$18.5\%$) 
where \safeclip greatly degrades the original model's knowledge, exposing a critical problem: fine-tuning for safety highly harnesses the original model's representations.

\begin{table*}[t]
    \centering
    \footnotesize
    \resizebox{\textwidth}{!}{
    \begin{tabular}{l|ccccc|cccccccccccc|c}
        \toprule
        \textbf{Method} & \textbf{ImageNet} & \textbf{V2} & \textbf{R} & \textbf{A} & \textbf{Sketch} & \textbf{CAL} & \textbf{PETS} & \textbf{FOOD} & \textbf{FLWR} & \textbf{C10} & \textbf{C100} &  \textbf{ESAT} & \textbf{CARS} & \textbf{AIR} & \textbf{DTD} & \textbf{SUN} & \textbf{UCF} & \textit{Mean} \\ \midrule
        \rowcolor{gray!20} CLIP~\cite{radford2021learning}  & 73.5 & 67.9 & 85.4 & 68.6 & 57.9 & 88.7 & 93.4 & 93.1 & 79.3 & 95.2 & 77.3  & 60.6 & 76.6 & 32.6 & 52.0 & 65.2 & 68.8 & 72.7 \\ \midrule
        \safeclip~\cite{poppi2024removing} & 56.1 & 49.2 & 67.7 & 34.4 & 36.2 & 79.3 & 78.7 & 78.2 & 50.7 & 88.8 & 63.9  & 27.6 & 44.6 & \textbf{16.6} & 41.1 & 54.5 & 54.6 & 54.2 \\ 
        I-P (\textit{Unsafe loss}) & 48.2 & 41.8 & 60.2 & 33.3 & 35.7 & 73.4 & 74.9 & 57.9 & 31.9 & 65.8 & 33.1 & 19.0 & 32.3 & 5.7 & 31.9 & 46.5 & 43.8 & 43.3 \\
        I-P (\textit{\safeclip loss}) & 60.4 & 54.4 & 71.2 & 49.4 & 39.6 & 85.8 & 81.9 & 79.1 & 46.5 & 83.3 & 52.1 & 37.0 & 47.2 & 11.9 & \textbf{42.2} & 57.0 & 55.6 & 56.3 \\
        \rowcolor{ModelLightBlue} \ours  & \textbf{62.3} & \textbf{56.5} & \textbf{79.8} & \textbf{57.2} & \textbf{48.9} & \textbf{86.6} & \textbf{82.2} & \textbf{85.8} & \textbf{52.3} & \textbf{91.1} & \textbf{69.5}  & \textbf{45.3} & \textbf{54.4} & 11.3 & 38.8 & \textbf{60.5} & \textbf{59.4} & \textbf{61.3} \\ 
        \bottomrule
    \end{tabular}
    }
    \vspace{-0.2cm}
    \caption{\textbf{Zero-shot Classification Accuracy on $17$ standard benchmarks.} The base CLIP model represents the upper bound (\colorbox{gray!20}{gray}). Among safety mitigation methods, \colorbox{ModelLightBlue}{\ours} achieves the highest accuracy, better preserving the model’s zero-shot capabilities.}
    \label{tab:zero_shot_performace}
    \vspace{-0.4cm}
\end{table*}

\ours enhances safety in both settings ($3.5\%$ on $\mathcal{T}_u\text{-}\mathcal{V}_{u+s}$ and $8.5\%$ on $\mathcal{V}_u\text{-}\mathcal{T}_{u+s}$), achieving the highest gain among training-free methods. However, it shows a decrease in zero-shot retrieval performance compared to the original CLIP model (\eg, $30\%$ on $\mathcal{T}_s\text{-}\mathcal{V}_s$ and $32.0\%$ on $\mathcal{V}_s\text{-}\mathcal{T}_s$). Therefore, we further evaluate its knowledge preservation in the first column, where \ours achieves the best zero-shot performance among all methods, including \safeclip. These results provide two key insights: (i) $\mathcal{T}_s\text{-}\mathcal{V}_s$ and $\mathcal{V}_s\text{-}\mathcal{T}_s$ alone are insufficient for evaluating knowledge preservation and (ii) \ours achieves the best balance between safety and knowledge preservation.  For a thorough evaluation of knowledge preservation, please refer to~\cref{sec:knowledge_preservation}. We now investigate safety using the safe preference metrics.

\para{Safe Preference Metrics} These metrics evaluate the model’s preference for a safe target across all possible queries.
The results of this evaluation are shown in the middle-right part of \cref{tab:visu_results}. These metrics confirm the unsafe nature of CLIP when tested on unsafe queries (\ie, $5.2\%$ on $\mathtt{P}_u^t$ and $7.6\%$ on $\mathtt{P}_u^v$), while achieving high safety on safe ones (\ie, $73.1\%$ on $\mathtt{P}_s^t$ and $87.4\%$ on $\mathtt{P}_s^v$).

The case of \textit{\baselineOne} highlights the need for novel metrics.
According to retrieval-based metric $\mathcal{T}_u\text{-}\mathcal{V}_{u+s}$, this method reduces safety for unsafe textual queries. However, by looking at $\mathtt{P}_u^t$ (\ie, preference of safe images when prompted with unsafe text), its safety improves. This is a direct example of the discussion in~\cref{sec:problem}. The drop in retrieval performance does not necessarily reflect lower safety but rather a degradation of the model’s knowledge, damaging its retrieval capabilities. This is further confirmed by the low zero-shot performance ($43.3\%$). Thus, our metrics successfully decouple safety from retrieval performance, enabling a more precise evaluation. Moreover, they expose the lower safety of \textit{\baselineOne} for safe queries ($\mathtt{P}_s^t$, $\mathtt{P}_s^v$).
In contrast, \textit{\baselineTwo} improves safety across all four metrics while preserving more knowledge, confirming the findings of the retrieval-based metrics: the method improves safety.

The second intriguing case 
is \safeclip. While outperforming all methods on unsafe queries (\ie, $19.0\%$ on $\mathtt{P}_u^t$, $34.1\%$ on $\mathtt{P}_u^v$), \textit{it degrades safety on safe ones} with a decrease of $23\%$ for text inputs ($\mathtt{P}_s^t$) and almost $6\%$ on $\mathtt{P}_u^t$. This shows the drawbacks of training-based methods.

\ours improves safety on unsafe inputs with $\text{+}6.5\%$ on $\mathtt{P}_u^t$ and $\text{+}12.9\%$ on $\mathtt{P}_u^v$, and outperforms \safeclip on safe queries (\eg, $\text{+}9.9\%$ on $\mathtt{P}_s^v$). These results confirm that \ours achieves the best balance between safety improvements and knowledge preservation.

\para{\textit{\bench} Metrics}\label{sec:model_safety_results}
We conclude by discussing the performance according to the \textit{\bench} metrics (last columns of \cref{tab:visu_results}) that combine the preferences to analyze safety across modalities and input type.
When analyzing the modalities ($\mathtt{Txt}_s$ and $\mathtt{Img}_s$), all methods struggle more with text queries, as the image modality score ($\mathtt{Img}_s$) is consistently lower than its textual counterpart ($\mathtt{Txt}_s$).
Interestingly, their gap tends to increase as models get safer: CLIP shows a $1.7\%$ difference, while \textit{\baselineTwo}, \ours, and \safeclip show increasing gaps of $2.7\%$, $8.3\%$, and $9.8\%$, respectively. This suggests that both training-free and training-based methods produce safer outputs for images compared to texts.
$\mathtt{PS}$ and $\mathtt{PU}$ (\ie, safety according to safe/unsafe input type), show consistent patterns for both baselines, with improvements for unsafe inputs ($\mathtt{PU}$) while lower performance on safe ones ($\mathtt{PS}$). Moreover, this latter case also applies to \safeclip, with a decrease of $21.6\%$, further confirming its unsafe performance on safe inputs. In contrast, \ours improves safety across both safe and unsafe inputs. Finally, all methods improve the group score $\mathtt{GS}$. 

Notably, these insights would have been more challenging to uncover with existing retrieval metrics, as they simultaneously capture retrieval and safety performance.

\subsubsection{Knowledge Preservation}
\label{sec:knowledge_preservation}
In this section, we evaluate the knowledge preservation of each method across $17$ classification tasks. We report the results in~\cref{tab:zero_shot_performace}, where CLIP serves as the upper bound, as it is the base model to which each method is applied. The results show that \textit{\baselineOne} consistently performs the worst on all tasks. The second-lowest performing model on average, \safeclip, reveals lower results than at least one competitor on all tasks except \textit{FGVCAircraft (AIR)}. This further confirms that the usability of \safeclip is compromised after its fine-tuning alignment. \textit{\baselineTwo} is the second-best performing method, with an average of $56.3\%$, demonstrating good knowledge preservation. 
\ours exhibits the best zero-shot performance, achieving the highest results across all tasks except two, with an average of $61.3\%$.

\begin{figure}[t]
    \centering
    \includegraphics[width=0.75\linewidth, trim=1.1cm 0 0 0.75cm, clip]{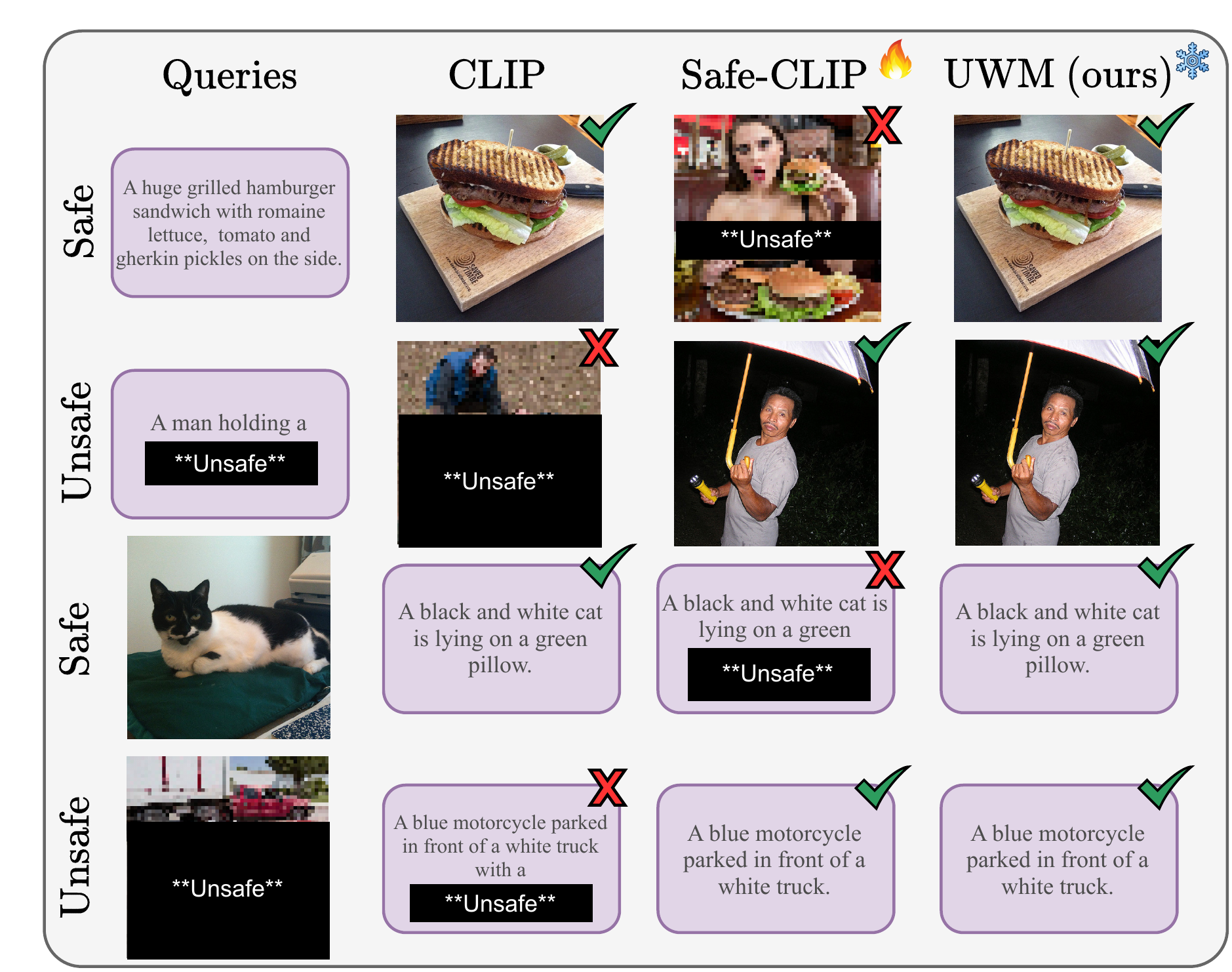}
    \vspace{-0.2cm}
    \caption{Qualitative results for $\mathcal{T}_u\text{-}\mathcal{V}_{u+s}$ and $\mathcal{V}_u\text{-}\mathcal{T}_{u+s}$ tasks.}
    \label{fig:qualitatives}
    \vspace{-0.5cm}
\end{figure}

\subsection{Analysis \& Discussion}\label{sec:ablation}
In this section, we show qualitative results of \ours and \safeclip, we ablate \ours's components, and we further test the proposed method on different VLM architectures. Finally, we apply \ours to LLaVA~\cite{liu2023llava} for captioning.

\para{Qualitatives} \cref{fig:qualitatives}~shows qualitative comparisons of CLIP, \safeclip, and \ours. CLIP retrieves unsafe content for unsafe queries in both modalities (second and fourth rows), while \safeclip consistently retrieves safe examples, demonstrating improved safety. Similarly, \ours mitigates CLIP’s unsafe behavior by retrieving safe content.
However, for \emph{safe queries} (first and second rows), \safeclip fails to retrieve safe content, confirming the significant failure mode exposed in the previous sections. In contrast, \ours preserves CLIP’s performance on safe data, demonstrating its robustness in this scenario. Additional qualitative results are provided in~\supp

\begin{table}[t]
    \centering
    \tiny
    \begin{tabular}{ccc|c|c@{\hspace{1.5em}}c@{\hspace{1.5em}}c@{\hspace{1.5em}}c@{\hspace{1.5em}}c}
        \toprule
        \multirow{2}{*}{\textbf{$\Phi^{\mathtt{uns}}$}} & \multirow{2}{*}{\textbf{$\frac{\Phi^{\mathtt{uns}}}{\Phi^{\mathtt{sf}}}$}} & \multirow{2}{*}{\textbf{Adapt}} & \textbf{Zero-Shot $(\uparrow)$} & \multicolumn{5}{c}{\textbf{\textit{\bench} Metrics $(\uparrow)$}} \\ \cmidrule{4-9}         
        & & & $\mathcal{V}_s$-$\mathcal{T}_s$ & $\mathtt{Txt}_\mathtt{s}$ & $\mathtt{Img}_\mathtt{s}$ & $\mathtt{PS}$ & $\mathtt{PU}$ & $\mathtt{GS}$ \\ \midrule
        \rowcolor{gray!20} -- & -- & -- & 39.8 & 6.4 & 4.7 & 67.5 & 1.7 & 1.2 \\ \midrule
        \checkmark & & & 24.3 & 18.3 & 14.0 & 60.9 & 5.9 & 4.6 \\ 
        \checkmark & \checkmark & & 16.2 & 37.7 & 22.6 & 61.0 & 15.7 & 13.0 \\ 
        \rowcolor{ModelLightBlue}  \checkmark & \checkmark & \checkmark & 32.0 & 19.1 & 10.8 & 67.8 & 5.5 & 4.5 \\ 
        \bottomrule
    \end{tabular}
    \vspace{-0.1cm}
    \caption{\textbf{Ablation study of \colorbox{ModelLightBlue}{\ours} across its components.} In \colorbox{gray!20}{gray} is the original version of CLIP.}
    \label{tab:ablation}
    \vspace{-0.2cm}
\end{table}

\begin{table}[]
    \centering
    \tiny
    \resizebox{0.9\linewidth}{!}{
    \begin{tabular}{c|cccc|ccccc}
        \toprule
        \multirow{2}{*}{\textbf{Model}} & \multicolumn{4}{c|}{\textbf{Basic Preference Metrics} $\left( \uparrow \right)$} & \multicolumn{5}{c}{\textbf{\textit{\bench} Metrics} $\left( \uparrow \right)$} \\ \cmidrule{2-10}
                                & $\mathtt{P}_s^t$ & $\mathtt{P}_u^t$ & $\mathtt{P}_s^v$ & $\mathtt{P}_u^v$ & $\mathtt{Txt}_\mathtt{s}$ & $\mathtt{Img}_\mathtt{s}$ & $\mathtt{PS}$ & $\mathtt{PU}$ & $\mathtt{GS}$ \\ \midrule
            ViT-L14~\cite{radford2021learning} & \textbf{73.1} & 5.2 & 87.4 & 7.6 & 6.4 & 4.7 & 67.5 & 1.7 & 1.2 \\ 
            \rowcolor{ModelLightBlue} \textit{+\ours} & 71.2  & \textbf{11.7} & \textbf{91.4} & \textbf{20.5} & \textbf{19.1} & \textbf{10.8} & \textbf{67.8} & \textbf{5.5} & \textbf{4.5} \\ \midrule
            ViT-B16~\cite{radford2021learning}  & \textbf{68.7} & 4.0 & 87.0 & 8.3 & 07.3 & 3.6 & \textbf{63.5} & 1.3 & 1.0  \\
            \rowcolor{ModelLightBlue} \textit{+\ours} & 67.1 & \textbf{8.2} & \textbf{89.1} & \textbf{16.4} & \textbf{14.8} & \textbf{7.3} & 62.7 & \textbf{3.0} & \textbf{2.0} \\  \midrule
            ViT-B32~\cite{radford2021learning}  & \textbf{67.4} & 4.6 & 86.9 & 9.2 & 8.1 & 4.1 & \textbf{62.2} & 1.6 & 1.1 \\
            \rowcolor{ModelLightBlue} \textit{+\ours} & 64.0 & \textbf{5.8} & \textbf{87.8} & \textbf{13.5} & \textbf{12.2} & \textbf{5.3} & 59.7 & \textbf{1.9} & \textbf{1.3} \\  \midrule
            CoCa~\cite{yu2022cocacontrastivecaptionersimagetext} & \textbf{79.5}  & 4.5 & 93.6 & 8.7 & 8.0 & 4.1 & \textbf{76.6} & 1.7 & 1.3 \\
            \rowcolor{ModelLightBlue} \textit{+\ours} & 77.6 & \textbf{5.6} & \textbf{94.9} & \textbf{15.5} & \textbf{15.4} & \textbf{5.2} & 74.8 & \textbf{2.6} & \textbf{2.2} \\  \midrule
            SigLIP~\cite{Zhai_2023_ICCV} & \textbf{73.6} & 3.5 & \textbf{92.8} & 7.9 & 7.2 & 3.1 & \textbf{70.7} & 1.5 & 1.0 \\
            \rowcolor{ModelLightBlue} \textit{+\ours} & 72.7 & \textbf{6.7} & 91.2 & \textbf{11.1} & \textbf{10.2} & \textbf{6.2} & 68.7 & \textbf{2.4} & \textbf{1.8} \\ \midrule
            Safe-CLIP~\cite{poppi2024removing} & 50.1 & \textbf{19.0} & 81.6 & 34.2 & 27.9 & \textbf{18.1} & 45.9 & 8.2 & 6.4 \\
            \rowcolor{ModelLightBlue} \textit{+\ours} & \textbf{50.6} & 18.5 & \textbf{86.8} & \textbf{42.2} & \textbf{37.7} & 17.3 & \textbf{47.4} & \textbf{9.0} & \textbf{7.5}  \\  
        \bottomrule
    \end{tabular} 
    }
    \vspace{-0.1cm}
    \caption{\textbf{Safety performance across architectures and pretraining strategies.}~\colorbox{ModelLightBlue}{\ours} reliably improves safety across models.}
    \label{tab:results_across_models} 
    \vspace{-0.5cm}
\end{table}

\para{Scoring Function} 
In~\cref{tab:ablation}, we ablate the scoring function for computing $\Phi$. We compare three variants: (i) using unsafe scores only $\Phi^{\mathtt{uns}}$, (ii) incorporating the aggregation score from~\cref{eq:aggregation_score}, and (iii) applying the adaptive selection of~\cref{eq:weights}. 
Relying solely on $\Phi^{\mathtt{uns}}$ degrades zero-shot performance (-$15.5$), but already improves safety (+$3.4$ $\mathtt{GS}$). Introducing the ratio $\Phi^{\mathtt{uns}}/\Phi^{\mathtt{sf}}$ further enhances safety (+$11.8$ $\mathtt{GS}$) but severely impacts zero-shot performance (-$23.6$), making the model unsuitable for downstream tasks. The best trade-off is achieved with adaptive selection, improving safety (+$4.5$ $\mathtt{GS}$) while minimizing zero-shot degradation (-$7.8$). Additional ablations are provided in the \supp

\para{Adaptability to Architectures} \ours is flexible and can be applied \textit{off-the-shelf} to any contrastive-based VLM. Accordingly, we extend our evaluation to different CLIP backbones~\cite{radford2021learning} and pretraining strategies, such as CoCa~\cite{yu2022cocacontrastivecaptionersimagetext} and SigLIP~\cite{Zhai_2023_ICCV}. Additionally, we evaluate its performance on \safeclip \cite{poppi2024removing}. We report the results in~\cref{tab:results_across_models}. \ours demonstrates consistent safety improvements across various CLIP backbones, enhancing safety on ViT-B16 (\eg, +$4.2$ $\mathtt{P}_u^t$, +$12.9$ $\mathtt{P}_u^v$ and +$12.7$ $\mathtt{Txt}_s$), while preserving its original behavior on safe queries (\eg, -$1.6$ $\mathtt{P}_s^t$). Similar improvements can be observed in ViT-B32 (\eg, +$4.3$ $\mathtt{P}_u^v$, +$12.9$ $\mathtt{P}_u^v$ and +$4.1$ $\mathtt{Txt}_s$). Moreover, \ours improves safety on models with different pre-training strategies, such as CoCa~\cite{yu2022cocacontrastivecaptionersimagetext} (\eg, +$6.8$ $\mathtt{P}_u^v$, +$7.4$ $\mathtt{Txt}_s$, and +$0.9$ $\mathtt{GS}$) and SigLIP~\cite{Zhai_2023_ICCV} (\eg, +$3.2$ $\mathtt{P}_u^t$, +$3.1$ $\mathtt{Img}_s$, and +$0.8$ $\mathtt{GS}$). Interestingly, our method further enhances \safeclip's safety (\eg, +$5.2$ $\mathtt{P}_s^v$, +$8$ $\mathtt{P}_u^v$, and +$1.1$ $\mathtt{GS}$). These results highlight the flexibility and effectiveness of \ours.

\para{LLaVA} We apply \ours to the vision encoder 
of LLaVA-1.5-13B~\cite{liu2023llava}. Following existing works~\cite{poppi2024removing}, we task LLaVA to caption unsafe images from the ViSU test set and report the percentage of Not Safe For Work (NSFW) generated content (measured with LLaMA-3-8B~\cite{grattafiori2024llama3herdmodels}) and toxicity score using the Perspective API~\cite{10.1145/3534678.3539147}. Additionally, we assess knowledge preservation using Rouge-L~\cite{lin-2004-rouge}, Bleu~\cite{papineni-etal-2002-bleu}, and Meteor~\cite{banerjee-lavie-2005-meteor}. We report the results in~\cref{tab:llava_results}. \ours effectively improves safety, reducing NSFW content by $9.8\%$ and toxicity by $3.4$, while preserving the original model's behavior (\eg, $0.01$ drop in Rouge-L). Moreover, after replacing LLaVA’s vision encoder with \safeclip~\cite{poppi2024removing}, we apply \ours to \safeclip within this setup. \ours further improves LLaVA safety by reducing NSFW generated content by $4.5\%$ and toxicity by $1\%$, while preserving its knowledge (\eg, equal Rouge-L and $0.01$ drop in Bleu). These results further validates \ours's applicability. 

\begin{table}[]
    \tiny
    \centering
    \resizebox{0.8\linewidth}{!}{
    \begin{tabular}{c|cc|ccc}
        \toprule
        \multirow{2}{*}{\textbf{Model}} & \multicolumn{2}{c|}{\textbf{Unsafety $\left ( \downarrow \right )$}} & \multicolumn{3}{c}{\textbf{Knowledge Preservation $\left ( \uparrow \right )$}} \\ \cmidrule{2-6}
        & \%NSFW & Toxicity & RougeL & Bleu & Meteor \\ \midrule
        LLaVA~\cite{liu2023llava} & 31.7 & 16.8 & \textbf{0.32} & \textbf{0.13} & \textbf{0.26} \\  
        \rowcolor{ModelLightBlue} \textit{+\ours} & \textbf{21.9} & \textbf{13.4} & 0.31 & 0.11 & 0.23 \\ \midrule
        \textit{+\safeclip} & 8.0 & 10.0 & \textbf{0.32} & \textbf{0.13} & \textbf{0.26} \\ 
        \rowcolor{ModelLightBlue} \textit{+\ours} & \textbf{3.5} & \textbf{9.0} & \textbf{0.32} & 0.12 & 0.24 \\ 
        \bottomrule 
    \end{tabular}
    }
    \vspace{-0.1cm}
    \caption{We apply \colorbox{ModelLightBlue}{\ours} to LLaVA\cite{liu2023llava} and LLaVA+\safeclip.}
    \label{tab:llava_results}
    \vspace{-0.6cm}
\end{table}

\section{Conclusion}\label{sec:conclusion}
This work investigates the safety challenges of Vision-Language Models (VLMs). We examine existing 
metrics~\cite{poppi2024removing} and find that they have limitations in assessing safety, as they rely on retrieval-based evaluations. Therefore, we complement them by introducing \textit{\bench}, a novel suite of metrics specifically designed for safety evaluation.
\textit{\bench} exposes a critical issue: training-based safety alignment techniques, such as \safeclip~\cite{poppi2024removing}, can compromise safe representations, leading to unsafe behaviors on safe queries. We propose \ours, a training-free method that identifies and manipulates unsafe weights in VLMs by analyzing how the information flow varies between safe and unsafe content. 
\ours achieves a better trade-off between safety and knowledge preservation, marking a promising first step toward training-free safety techniques for VLMs. Finally, we hope \bench will serve as a valuable resource for the community, enabling researchers to uncover unsafe behaviors in VLMs.

\para{Limitations} While \ours is applicable to various architectures and VLMs (\cref{tab:results_across_models} and~\cref{tab:llava_results}), it relies on weight localization and manipulation. The complexity of large-scale VLMs challenges the isolation of unsafe weights without affecting model capabilities, as individual parameters may encode overlapping knowledge. As both \ours and \safeclip~\cite{poppi2024removing} exhibit limitations, VLM safety remains an open challenge. Lastly, we focus mainly on contrastive-based VLMs, leaving other architectures for future research.

\para{Acknowledgments} This work was supported by the EU Horizon project “ELIAS - European Lighthouse of AI for Sustainability” (No. 101120237), MUR PNRR project FAIR (PE00000013) funded by the NextGenerationEU, and the FIS project GUIDANCE (Debugging Computer Vision Models via Controlled Cross-modal Generation) (No. FIS2023-03251).

\clearpage
{
    \small
    \bibliographystyle{ieeenat_fullname}
    \bibliography{main}

@String(CVPR= {IEEE Conf. Comput. Vis. Pattern Recog.})

@String(ICCV= {Int. Conf. Comput. Vis.})

@String(ECCV= {Eur. Conf. Comput. Vis.})

@String(NIPS= {Adv. Neural Inform. Process. Syst.})

@String(ICLR = {Int. Conf. Learn. Represent.})

@String(CVPR  = {CVPR})

@String(ICCV  = {ICCV})

@String(ECCV  = {ECCV})

@String(NIPS  = {NeurIPS})

@String(ICLR  = {ICLR})

@inproceedings{radford2021learning,
  title={Learning transferable visual models from natural language supervision},
  author={Radford, Alec and Kim, Jong Wook and Hallacy, Chris and Ramesh, Aditya and Goh, Gabriel and Agarwal, Sandhini and Sastry, Girish and Askell, Amanda and Mishkin, Pamela and Clark, Jack and others},
  booktitle={ICML},
  year={2021}
}

@inproceedings{poppi2024removing,
  title={{Safe-CLIP: Removing NSFW Concepts from Vision-and-Language Models}},
  author={Poppi, Samuele and Poppi, Tobia and Cocchi, Federico and Cornia, Marcella and Baraldi, Lorenzo and Cucchiara, Rita},
  booktitle=ECCV,
  year={2024}
}

@article{birhane2021multimodal,
  title={Multimodal datasets: misogyny, pornography, and malignant stereotypes},
  author={Birhane, Abeba and Prabhu, Vinay Uday and Kahembwe, Emmanuel},
  journal={arXiv preprint arXiv:2110.01963},
  year={2021}
}

@inproceedings{banerjee-lavie-2005-meteor,
    title = "{METEOR}: An Automatic Metric for {MT} Evaluation with Improved Correlation with Human Judgments",
    author = "Banerjee, Satanjeev  and Lavie, Alon",
    booktitle = "Proceedings of the {ACL} Workshop on Intrinsic and Extrinsic Evaluation Measures for Machine Translation and/or Summarization",
    year = "2005",
}

@misc{grattafiori2024llama3herdmodels,
      title={The Llama 3 Herd of Models}, 
      author ={Llama Team},
      year={2024},
      eprint={2407.21783},
      archivePrefix={arXiv},
      url={https://arxiv.org/abs/2407.21783},
}

@inproceedings{10.1145/3534678.3539147,
author = {Lees, Alyssa and Tran, Vinh Q. and Tay, Yi and Sorensen, Jeffrey and Gupta, Jai and Metzler, Donald and Vasserman, Lucy},
title = {A New Generation of Perspective API: Efficient Multilingual Character-level Transformers},
year = {2022},
booktitle = {Proceedings of the 28th ACM SIGKDD Conference on Knowledge Discovery and Data Mining}
}

@inproceedings{papineni-etal-2002-bleu,
    title = "{B}leu: a Method for Automatic Evaluation of Machine Translation",
    author = "Papineni, Kishore  and Roukos, Salim  and Ward, Todd  and Zhu, Wei-Jing",
    booktitle = "Proceedings of the 40th Annual Meeting of the Association for Computational Linguistics",
    year = "2002",
}

@inproceedings{lin-2004-rouge,
    title = "{ROUGE}: A Package for Automatic Evaluation of Summaries",
    author = "Lin, Chin-Yew",
    booktitle = "Text Summarization Branches Out",
    year = "2004",
}

@InProceedings{Schramowski_2023_CVPR,
    author    = {Schramowski, Patrick and Brack, Manuel and Deiseroth, Bj\"orn and Kersting, Kristian},
    title     = {Safe Latent Diffusion: Mitigating Inappropriate Degeneration in Diffusion Models},
    booktitle = CVPR,
    year      = {2023}
}

@article{wang2023modelediting,
  title={Knowledge editing for large language models: A survey},
  author={Wang, Song and Zhu, Yaochen and Liu, Haochen and Zheng, Zaiyi and Chen, Chen and Li, Jundong},
  journal={ACM Computing Surveys},
  year={2023}
}

@inproceedings{mitchellfast,
  title={Fast Model Editing at Scale},
  author={Mitchell, Eric and Lin, Charles and Bosselut, Antoine and Finn, Chelsea and Manning, Christopher D},
  booktitle={ICLR},
  year={2022}
}

@inproceedings{mitchell2022memory,
  title={Memory-based model editing at scale},
  author={Mitchell, Eric and Lin, Charles and Bosselut, Antoine and Manning, Christopher D and Finn, Chelsea},
  booktitle={ICML},
  year={2022}
}

@inproceedings{gandikota2023erasing,
  title={Erasing concepts from diffusion models},
  author={Gandikota, Rohit and Materzynska, Joanna and Fiotto-Kaufman, Jaden and Bau, David},
  booktitle=ICCV,
  year={2023}
}

@inproceedings{
    yuan2024gpt,
    title={{GPT}-4 Is Too Smart To Be Safe: Stealthy Chat with {LLM}s via Cipher},
    author={Youliang Yuan and Wenxiang Jiao and Wenxuan Wang and Jen-tse Huang and Pinjia He and Shuming Shi and Zhaopeng Tu},
    booktitle=ICLR,
    year={2024}
}

@inproceedings{
    liu2024autodan,
    title={Auto{DAN}: Generating Stealthy Jailbreak Prompts on Aligned Large Language Models},
    author={Xiaogeng Liu and Nan Xu and Muhao Chen and Chaowei Xiao},
    booktitle=ICLR,
    year={2024}
}

@inproceedings{shen2024anything,
  title={" do anything now": Characterizing and evaluating in-the-wild jailbreak prompts on large language models},
  author={Shen, Xinyue and Chen, Zeyuan and Backes, Michael and Shen, Yun and Zhang, Yang},
  booktitle={Proceedings of the 2024 on ACM SIGSAC Conference on Computer and Communications Security},
  year={2024}
}

@inproceedings{mehrotra2025tree,
  title={Tree of attacks: Jailbreaking black-box llms automatically},
  author={Mehrotra, Anay and Zampetakis, Manolis and Kassianik, Paul and Nelson, Blaine and Anderson, Hyrum and Singer, Yaron and Karbasi, Amin},
  booktitle=NIPS,
  year={2024}
}

@inproceedings{NEURIPS2023_44956951,
 author = {Ma, Xinyin and Fang, Gongfan and Wang, Xinchao},
 booktitle = NIPS,
 title = {LLM-Pruner: On the Structural Pruning of Large Language Models},
 year = {2023}
}

@inproceedings{
    xu2024besa,
    title={{BESA}: Pruning Large Language Models with Blockwise Parameter-Efficient Sparsity Allocation},
    author={Peng Xu and Wenqi Shao and Mengzhao Chen and Shitao Tang and Kaipeng Zhang and Peng Gao and Fengwei An and Yu Qiao and Ping Luo},
    booktitle=ICLR,
    year={2024}
}

@inproceedings{
    hsu2024safe,
    title={Safe Lo{RA}: The Silver Lining of Reducing Safety Risks when Finetuning Large Language Models},
    author={Chia-Yi Hsu and Yu-Lin Tsai and Chih-Hsun Lin and Pin-Yu Chen and Chia-Mu Yu and Chun-Ying Huang},
    booktitle=NIPS,
    year={2024}
}

@article{das2023beyond,
  title={Beyond size: How gradients shape pruning decisions in large language models},
  author={Das, Rocktim Jyoti and Sun, Mingjie and Ma, Liqun and Shen, Zhiqiang},
  journal={arXiv preprint arXiv:2311.04902},
  year={2023}
}

@article{wang2019learning,
  title={Learning robust global representations by penalizing local predictive power},
  author={Wang, Haohan and Ge, Songwei and Lipton, Zachary and Xing, Eric P},
  journal={Advances in neural information processing systems},
  year={2019}
}

@inproceedings{recht2019imagenet,
  title={Do imagenet classifiers generalize to imagenet?},
  author={Recht, Benjamin and Roelofs, Rebecca and Schmidt, Ludwig and Shankar, Vaishaal},
  booktitle={International conference on machine learning},
  year={2019}
}

@inproceedings{hendrycks2021many,
  title={The many faces of robustness: A critical analysis of out-of-distribution generalization},
  author={Hendrycks, Dan and Basart, Steven and Mu, Norman and Kadavath, Saurav and Wang, Frank and Dorundo, Evan and Desai, Rahul and Zhu, Tyler and Parajuli, Samyak and Guo, Mike and others},
  booktitle={Proceedings of the IEEE/CVF international conference on computer vision},
  year={2021}
}

@inproceedings{hendrycks2021natural,
  title={Natural adversarial examples},
  author={Hendrycks, Dan and Zhao, Kevin and Basart, Steven and Steinhardt, Jacob and Song, Dawn},
  booktitle={Proceedings of the IEEE/CVF conference on computer vision and pattern recognition},
  year={2021}
}

@inproceedings{liu2023llava,
    author      = {Liu, Haotian and Li, Chunyuan and Wu, Qingyang and Lee, Yong Jae},
    title       = {Visual Instruction Tuning},
    booktitle   = NIPS,
    year        = {2023}
  }

@InProceedings{Gandikota_2024_WACV,
    author    = {Gandikota, Rohit and Orgad, Hadas and Belinkov, Yonatan and Materzy\'nska, Joanna and Bau, David},
    title     = {Unified Concept Editing in Diffusion Models},
    booktitle = {WACV},
    year      = {2024},
}

@InProceedings{zhang2024generate,
  title={To generate or not? safety-driven unlearned diffusion models are still easy to generate unsafe images... for now},
  author={Zhang, Yimeng and Jia, Jinghan and Chen, Xin and Chen, Aochuan and Zhang, Yihua and Liu, Jiancheng and Ding, Ke and Liu, Sijia},
  booktitle=ECCV,
  year={2024}
}

@inproceedings{liu2025latent,
  title={Latent guard: a safety framework for text-to-image generation},
  author={Liu, Runtao and Khakzar, Ashkan and Gu, Jindong and Chen, Qifeng and Torr, Philip and Pizzati, Fabio},
  booktitle=ECCV,
  year={2025}
}

@InProceedings{liu2023cones,
  title={Cones: Concept neurons in diffusion models for customized generation},
  author={Liu, Zhiheng and Feng, Ruili and Zhu, Kai and Zhang, Yifei and Zheng, Kecheng and Liu, Yu and Zhao, Deli and Zhou, Jingren and Cao, Yang},
  booktitle={ICML},
  year={2023}
}

@inproceedings{ringabell,
    title={Ring-A-Bell! How Reliable are Concept Removal Methods For Diffusion Models?},
    author={Yu-Lin Tsai and Chia-Yi Hsu and Chulin Xie and Chih-Hsun Lin and Jia-You Chen and Bo Li and Pin-Yu Chen and Chia-Mu Yu and Chun-Ying Huang},
    booktitle=ICLR,
    year={2024}
}

@InProceedings{Farina_2024_CVPR,
    author    = {Farina, Matteo and Mancini, Massimiliano and Cunegatti, Elia and Liu, Gaowen and Iacca, Giovanni and Ricci, Elisa},
    title     = {MULTIFLOW: Shifting Towards Task-Agnostic Vision-Language Pruning},
    booktitle = CVPR,
    year      = {2024}
}

@inproceedings{deng2009imagenet,
  title={Imagenet: A large-scale hierarchical image database},
  author={Deng, Jia and Dong, Wei and Socher, Richard and Li, Li-Jia and Li, Kai and Fei-Fei, Li},
  booktitle=CVPR,
  year={2009}
}

@inproceedings{nilsback2008automated,
  title={Automated flower classification over a large number of classes},
  author={Nilsback, Maria-Elena and Zisserman, Andrew},
  booktitle={2008 Sixth Indian conference on computer vision, graphics \& image processing},
  year={2008}
}

@inproceedings{cimpoi2014describing,
  title={Describing textures in the wild},
  author={Cimpoi, Mircea and Maji, Subhransu and Kokkinos, Iasonas and Mohamed, Sammy and Vedaldi, Andrea},
  booktitle=CVPR,
  year={2014}
}

@inproceedings{parkhi2012cats,
  title={Cats and dogs},
  author={Parkhi, Omkar M and Vedaldi, Andrea and Zisserman, Andrew and Jawahar, CV},
  booktitle=CVPR,
  year={2012}
}

@inproceedings{krause20133d,
  title={3d object representations for fine-grained categorization},
  author={Krause, Jonathan and Stark, Michael and Deng, Jia and Fei-Fei, Li},
  booktitle={ICCV-WS},
  year={2013}
}

@article{soomro2012ucf101,
  title={UCF101: A dataset of 101 human actions classes from videos in the wild},
  author={Soomro, K},
  journal={arXiv preprint arXiv:1212.0402},
  year={2012}
}

@inproceedings{fei2004learning,
  title={Learning generative visual models from few training examples: An incremental bayesian approach tested on 101 object categories},
  author={Fei-Fei, Li and Fergus, Rob and Perona, Pietro},
  booktitle={CVPR-WS},
  year={2004}
}

@inproceedings{bossard2014food,
  title={Food-101--mining discriminative components with random forests},
  author={Bossard, Lukas and Guillaumin, Matthieu and Van Gool, Luc},
  booktitle=ECCV,
  year={2014}
}

@inproceedings{xiao2010sun,
  title={Sun database: Large-scale scene recognition from abbey to zoo},
  author={Xiao, Jianxiong and Hays, James and Ehinger, Krista A and Oliva, Aude and Torralba, Antonio},
  booktitle={CVPR},
  year={2010}
}

@article{maji2013fine,
  title={Fine-grained visual classification of aircraft},
  author={Maji, Subhransu and Rahtu, Esa and Kannala, Juho and Blaschko, Matthew and Vedaldi, Andrea},
  journal={arXiv preprint arXiv:1306.5151},
  year={2013}
}

@article{helber2019eurosat,
  title={Eurosat: A novel dataset and deep learning benchmark for land use and land cover classification},
  author={Helber, Patrick and Bischke, Benjamin and Dengel, Andreas and Borth, Damian},
  journal={IEEE Journal of Selected Topics in Applied Earth Observations and Remote Sensing},
  year={2019}
}

@article{krizhevsky2009learning,
  title={Learning multiple layers of features from tiny images},
  author={Krizhevsky, Alex and Hinton, Geoffrey and others},
  year={2009},
  publisher={Toronto, ON, Canada}
}

@InProceedings{Thrush_2022_CVPR,
    author    = {Thrush, Tristan and Jiang, Ryan and Bartolo, Max and Singh, Amanpreet and Williams, Adina and Kiela, Douwe and Ross, Candace},
    title     = {Winoground: Probing Vision and Language Models for Visio-Linguistic Compositionality},
    booktitle = CVPR,
    month     = {June},
    year      = {2022}
}

@misc{chakraborty2024crossmodalsafetyalignmenttextual,
      title={Cross-Modal Safety Alignment: Is textual unlearning all you need?}, 
      author={Trishna Chakraborty and Erfan Shayegani and Zikui Cai and Nael Abu-Ghazaleh and M. Salman Asif and Yue Dong and Amit K. Roy-Chowdhury and Chengyu Song},
      year={2024},
      eprint={2406.02575},
      archivePrefix={arXiv},
}

@article{yuan2024refuse,
  title={Refuse whenever you feel unsafe: Improving safety in llms via decoupled refusal training},
  author={Yuan, Youliang and Jiao, Wenxiang and Wang, Wenxuan and Huang, Jen-tse and Xu, Jiahao and Liang, Tian and He, Pinjia and Tu, Zhaopeng},
  journal={arXiv preprint arXiv:2407.09121},
  year={2024}
}

@InProceedings{Zhai_2023_ICCV,
    author    = {Zhai, Xiaohua and Mustafa, Basil and Kolesnikov, Alexander and Beyer, Lucas},
    title     = {Sigmoid Loss for Language Image Pre-Training},
    booktitle = ICCV,
    month     = {October},
    year      = {2023}
}

@misc{yu2022cocacontrastivecaptionersimagetext,
      title={CoCa: Contrastive Captioners are Image-Text Foundation Models}, 
      author={Jiahui Yu and Zirui Wang and Vijay Vasudevan and Legg Yeung and Mojtaba Seyedhosseini and Yonghui Wu},
      year={2022},
      eprint={2205.01917},
      archivePrefix={arXiv}
}

@inproceedings{shi2023upop,
  title={Upop: Unified and progressive pruning for compressing vision-language transformers},
  author={Shi, Dachuan and Tao, Chaofan and Jin, Ying and Yang, Zhendong and Yuan, Chun and Wang, Jiaqi},
  booktitle={ICML},
  year={2023}
}

@article{sun2023simple,
  title={A simple and effective pruning approach for large language models},
  author={Sun, Mingjie and Liu, Zhuang and Bair, Anna and Kolter, J Zico},
  journal={arXiv preprint arXiv:2306.11695},
  year={2023}
}

@article{li2024red,
  title={Red teaming visual language models},
  author={Li, Mukai and Li, Lei and Yin, Yuwei and Ahmed, Masood and Liu, Zhenguang and Liu, Qi},
  journal={arXiv preprint arXiv:2401.12915},
  year={2024}
}

@article{perez2022red,
  title={Red teaming language models with language models},
  author={Perez, Ethan and Huang, Saffron and Song, Francis and Cai, Trevor and Ring, Roman and Aslanides, John and Glaese, Amelia and McAleese, Nat and Irving, Geoffrey},
  journal={arXiv preprint arXiv:2202.03286},
  year={2022}
}

@article{zou2023universal,
  title={Universal and transferable adversarial attacks on aligned language models},
  author={Zou, Andy and Wang, Zifan and Carlini, Nicholas and Nasr, Milad and Kolter, J Zico and Fredrikson, Matt},
  journal={arXiv preprint arXiv:2307.15043},
  year={2023}
}

@article{gallegos2024bias,
  title={Bias and fairness in large language models: A survey},
  author={Gallegos, Isabel O and Rossi, Ryan A and Barrow, Joe and Tanjim, Md Mehrab and Kim, Sungchul and Dernoncourt, Franck and Yu, Tong and Zhang, Ruiyi and Ahmed, Nesreen K},
  journal={Computational Linguistics},
  pages={1--79},
  year={2024},
  publisher={MIT Press 255 Main Street, 9th Floor, Cambridge, Massachusetts 02142, USA~…}
}

@article{wei2024jailbroken,
  title={Jailbroken: How does llm safety training fail?},
  author={Wei, Alexander and Haghtalab, Nika and Steinhardt, Jacob},
  journal={NeurIPS},
  year={2024}
}

@article{bommasani2021opportunities,
  title={On the opportunities and risks of foundation models},
  author={Bommasani, Rishi and Hudson, Drew A and Adeli, Ehsan and Altman, Russ and Arora, Simran and von Arx, Sydney and Bernstein, Michael S and Bohg, Jeannette and Bosselut, Antoine and Brunskill, Emma and others},
  journal={arXiv preprint arXiv:2108.07258},
  year={2021}
}

@article{brown2020language,
  title={Language models are few-shot learners},
  author={Brown, Tom B},
  journal={arXiv preprint arXiv:2005.14165},
  year={2020}
}

@article{touvron2023llama,
  title={Llama: Open and efficient foundation language models},
  author={Touvron, Hugo and Lavril, Thibaut and Izacard, Gautier and Martinet, Xavier and Lachaux, Marie-Anne and Lacroix, Timoth{\'e}e and Rozi{\`e}re, Baptiste and Goyal, Naman and Hambro, Eric and Azhar, Faisal and others},
  journal={arXiv preprint arXiv:2302.13971},
  year={2023}
}

@inproceedings{d2024openbias,
  title={OpenBias: Open-set Bias Detection in Text-to-Image Generative Models},
  author={D'Inc{\`a}, Moreno and Peruzzo, Elia and Mancini, Massimiliano and Xu, Dejia and Goel, Vidit and Xu, Xingqian and Wang, Zhangyang and Shi, Humphrey and Sebe, Nicu},
  booktitle={CVPR},
  year={2024}
}

@inproceedings{rombach2021high,
  title={High-resolution image synthesis with latent diffusion models. 2022 IEEE},
  author={Rombach, Robin and Blattmann, Andreas and Lorenz, Dominik and Esser, Patrick and Ommer, Bj{\"o}rn},
  booktitle={CVPR},
  year={2021}
}

@article{ramesh2022hierarchical,
  title={Hierarchical text-conditional image generation with clip latents},
  author={Ramesh, Aditya and Dhariwal, Prafulla and Nichol, Alex and Chu, Casey and Chen, Mark},
  journal={arXiv preprint arXiv:2204.06125},
  volume={1},
  number={2},
  pages={3},
  year={2022}
}

@article{meng2022mass,
  title={Mass-editing memory in a transformer},
  author={Meng, Kevin and Sharma, Arnab Sen and Andonian, Alex and Belinkov, Yonatan and Bau, David},
  journal={arXiv preprint arXiv:2210.07229},
  year={2022}
}

@article{meng2022locating,
  title={Locating and editing factual associations in GPT},
  author={Meng, Kevin and Bau, David and Andonian, Alex and Belinkov, Yonatan},
  journal={NeurIPS},
  year={2022}
}

@article{arad2023refact,
  title={Refact: Updating text-to-image models by editing the text encoder},
  author={Arad, Dana and Orgad, Hadas and Belinkov, Yonatan},
  journal={arXiv preprint arXiv:2306.00738},
  year={2023}
}

@inproceedings{orgad2023editing,
  title={Editing implicit assumptions in text-to-image diffusion models},
  author={Orgad, Hadas and Kawar, Bahjat and Belinkov, Yonatan},
  booktitle={ICCV},
  year={2023}
}

@article{alizadeh2022prospect,
  title={Prospect pruning: Finding trainable weights at initialization using meta-gradients},
  author={Alizadeh, Milad and Tailor, Shyam A and Zintgraf, Luisa M and van Amersfoort, Joost and Farquhar, Sebastian and Lane, Nicholas Donald and Gal, Yarin},
  journal={arXiv preprint arXiv:2202.08132},
  year={2022}
}

@article{wang2020picking,
  title={Picking winning tickets before training by preserving gradient flow},
  author={Wang, Chaoqi and Zhang, Guodong and Grosse, Roger},
  journal={arXiv preprint arXiv:2002.07376},
  year={2020}
}

@article{lee2018snip,
  title={Snip: Single-shot network pruning based on connection sensitivity},
  author={Lee, Namhoon and Ajanthan, Thalaiyasingam and Torr, Philip HS},
  journal={arXiv preprint arXiv:1810.02340},
  year={2018}
}

@article{dong2017learning,
  title={Learning to prune deep neural networks via layer-wise optimal brain surgeon},
  author={Dong, Xin and Chen, Shangyu and Pan, Sinno},
  journal={NeurIPS},
  year={2017}
}

@article{han2015deep,
  title={Deep compression: Compressing deep neural networks with pruning, trained quantization and huffman coding},
  author={Han, Song and Mao, Huizi and Dally, William J},
  journal={arXiv preprint arXiv:1510.00149},
  year={2015}
}

@article{lee2020layer,
  title={Layer-adaptive sparsity for the magnitude-based pruning},
  author={Lee, Jaeho and Park, Sejun and Mo, Sangwoo and Ahn, Sungsoo and Shin, Jinwoo},
  journal={arXiv preprint arXiv:2010.07611},
  year={2020}
}

@inproceedings{prabhu2021large,
  title={Large datasets: A pyrrhic win for computer vision},
  author={Prabhu, Vinay Uday and Birhane, Abeba},
  booktitle={Institute of Electrical and Electronics Engineers/Computer Vision Foundation Conference on Applications of Computer Vision},
  year={2021}
}

@inproceedings{schramowski2022can,
  title={Can machines help us answering question 16 in datasheets, and in turn reflecting on inappropriate content?},
  author={Schramowski, Patrick and Tauchmann, Christopher and Kersting, Kristian},
  booktitle={Proceedings of the 2022 ACM Conference on Fairness, Accountability, and Transparency},
  pages={1350--1361},
  year={2022}
}

@inproceedings{goodfellow2013empirical,
  title={An Empirical Investigation of Catastrophic Forgetting in Gradient-Based Neural Networks},
  author={Goodfellow, Ian J and Mirza, Mehdi and Xiao, Da and Courville, Aaron and Bengio, Yoshua},
  booktitle={ICLR},
  year={2014}
}

@article{li2017learning,
  title={Learning without forgetting},
  author={Li, Zhizhong and Hoiem, Derek},
  journal={IEEE T-PAMI},
  volume={40},
  number={12},
  pages={2935--2947},
  year={2017},
  publisher={IEEE}
}
}
\clearpage
\maketitlesupplementary
\appendix 

In this supplementary material, we provide additional ablations and analysis. Specifically, in~\cref{supp_sec:ablation} we i) provide a comprehensive ablation across all components of \ours and ii) expand baseline descriptions. Finally,~\cref{supp_sec:qualitatives} shows further qualitative results and failure cases.

\begin{table*}[!ht]
    \footnotesize
    \centering
    \begin{tabular}{cccc|c|c@{\hspace{1.5em}}c@{\hspace{1.5em}}c@{\hspace{1.5em}}c@{\hspace{1.5em}}c}
        \toprule
        \multirow{2}{*}{\textbf{$\Phi^{\mathtt{uns}}$}} & \multirow{2}{*}{\textbf{$\frac{\Phi^{\mathtt{uns}}}{\Phi^{\mathtt{sf}}}$}} & \multirow{2}{*}{\textbf{$\Phi^{\mathtt{uns}}-\Phi^{\mathtt{sf}}$}} & \multirow{2}{*}{\textbf{Adapt}}& \textbf{Zero-Shot $(\uparrow)$} & \multicolumn{5}{c}{\textbf{\bench Metrics $(\uparrow)$}} \\ \cmidrule{5-10}         

        & & & & $\mathcal{V}_s$-$\mathcal{T}_s$ & $\mathtt{Txt}_\mathtt{s}$ & $\mathtt{Img}_\mathtt{s}$ & $\mathtt{PS}$ & $\mathtt{PU}$ & $\mathtt{GS}$ \\ \midrule
        \rowcolor{gray!20} -- & -- & -- & -- & 39.8 & 6.4 & 4.7 & 67.5 & 1.7 & 1.2 \\ \midrule
        \checkmark & & & & 24.3 & 18.3 & 14.0 & 60.9 & 5.9 & 4.6 \\ 
        \checkmark & & &\checkmark & 30.6 & 15.9 & 9.6 & 64.3 & 4.3 & 3.4 \\ 
        \checkmark & & \checkmark & & 31.0 & 15.0 & 8.5 & 66.2 & 3.8 & 2.6 \\ 
        \checkmark & & \checkmark & \checkmark & 38.7 & 8.6 & 4.4 & 66.9 & 2.0 & 1.3 \\ 
        \checkmark & \checkmark & & & 16.2 & 37.7 & 22.6 & 61.0 & 15.7 & 13.0 \\ 
        \rowcolor{ModelLightBlue}  \checkmark & \checkmark  & & \checkmark & 32.0 & 19.1 & 10.9 & 67.8 & 5.5 & 4.5 \\ 
        \bottomrule
    \end{tabular}
    \caption{We further ablate \colorbox{ModelLightBlue}{\ours} across its components. In \colorbox{gray!20}{gray} the original version of CLIP.}
    \label{tab:supp_ablation}
    \vspace{-0.2cm}
\end{table*}

\begin{table}[t]
    \centering
    \tiny
    \begin{tabular}{c|c@{\hspace{1em}}c|c|c@{\hspace{1.5em}}c@{\hspace{1.5em}}c@{\hspace{1.5em}}c@{\hspace{1.5em}}c}
        \toprule
        \multirow{2}{*}{\textbf{Method}}  & \multicolumn{2}{c|}{\textbf{Priors}} & \textbf{Zero-Shot ($\uparrow$)} & \multicolumn{5}{c}{\textbf{\bench Metrics ($\uparrow$)}} \\ \cmidrule{2-9}
        & $\left | W_\mathcal{T} \right |$ & $\left | W_\mathcal{V} \right |$ & $\mathcal{V}_s$-$\mathcal{T}_s$ & $\mathtt{Txt}_\mathtt{s}$ & $\mathtt{Img}_\mathtt{s}$ & $\mathtt{PS}$ & $\mathtt{PU}$ & $\mathtt{GS}$ \\ \midrule
        \rowcolor{gray!20} CLIP   & - & - & 39.8 & 6.4 & 4.7 & 67.5 & 1.7 & 1.2 \\ \midrule
        \multirow{4}{*}{\ours}  & & & 38.0 & 10.8 & 5.7 & \textbf{70.1} & 2.5 & 1.9 \\
         & \checkmark & & 32.0 & \textbf{19.1} & \textbf{10.8} & 67.8 & \textbf{5.5} & \textbf{4.5} \\
         & & \checkmark & 0.0 & -- & -- & -- & -- & --\\
         & \checkmark & \checkmark & 0.0 & -- & -- & -- & -- & --\\
        \bottomrule
    \end{tabular}
    \caption{We further ablate \ours's scoring function.}
    \label{tab:prior_ablation}
\end{table}

\begin{table*}[!htbp]
    \footnotesize
    \centering
    \begin{tabular}{cccc|cccc|c|c@{\hspace{1.5em}}c@{\hspace{1.5em}}c@{\hspace{1.5em}}c@{\hspace{1.5em}}c}
        \toprule
        \multicolumn{4}{c|}{\textbf{Text Encoder}} & \multicolumn{4}{c|}{\textbf{Image Encoder}} & \multirow{2}{*}{\textbf{Zero-Shot $(\uparrow)$}} & \multicolumn{5}{c}{\multirow{2}{*}{\textbf{\bench Metrics $(\uparrow)$}}} \\   

        \multicolumn{4}{c|}{\textbf{Layers}} & \multicolumn{4}{c|}{\textbf{Layers}} & & & & & & \\ \midrule
        Fc1 & Fc2 & Value & Out & Fc1 & Fc2 & Value & Out & $\mathcal{V}_s$-$\mathcal{T}_s$ & $\mathtt{Txt}_\mathtt{s}$ & $\mathtt{Img}_\mathtt{s}$ & $\mathtt{PS}$ & $\mathtt{PU}$ & $\mathtt{GS}$ \\ \midrule
        \rowcolor{gray!20} -- & -- & -- & -- & -- & -- & -- & -- & 39.8 & 6.4 & 4.7 & 67.5 & 1.7 & 1.2 \\ \midrule
        
        \checkmark & & & & \checkmark & & & & 0.0 & -- & -- & -- & -- & -- \\ 
        \checkmark & & & & & \checkmark & & & 2.7 & -- & -- & -- & -- & -- \\ 
        \checkmark & & & & & & \checkmark & & 2.6 & -- & -- & -- & -- & -- \\ 
        \checkmark & & & & & & & \checkmark & 3.1 & -- & -- & -- & -- & -- \\ \midrule
        & \checkmark & & & \checkmark & & & & 0.0 & -- & -- & -- & -- & -- \\ 
        & \checkmark & & & & \checkmark & & & 1.3 & -- & -- & -- & -- & -- \\ 
        & \checkmark & & & & & \checkmark & & 1.2 & -- & -- & -- & -- & -- \\ 
        & \checkmark & & & & & & \checkmark & 1.3 & -- & -- & -- & -- & -- \\ \midrule
        & & \checkmark & & \checkmark & & & & 0.0 & -- & -- & -- & -- & -- \\ 
        & & \checkmark & & & \checkmark & & & 23.0 & 29.3 & 5.0 & 56.3 & 4.5 & 2.7 \\ 
        & & \checkmark & & & & \checkmark & & 20.6 & 30.4 & 4.6 & 49.8 & 3.8 & 2.4 \\ 
        & & \checkmark & & & & & \checkmark & 24.2 & 29.8 & 3.8 & 49.8 & 3.6 & 2.0 \\ \midrule
        & & & \checkmark & \checkmark & & & & 0.0 & -- & -- & -- & -- & -- \\ 
        \rowcolor{ModelLightBlue} & & & \checkmark & & \checkmark & & & 32.0 & 19.1 & 10.9 & 67.8 & 5.5 & 4.5 \\ 
        & & & \checkmark & & & \checkmark & & 28.8 & 19.0 & 12.1 & 65.1 & 5.8 & 4.6 \\ 
        & & & \checkmark & & & & \checkmark & 32.4 & 18.7 & 9.7 & 66.3 & 5.0 & 4.2 \\ 
        \bottomrule
    \end{tabular}
    \caption{We apply \colorbox{ModelLightBlue}{\ours} to different layers. In \colorbox{gray!20}{gray} the original version of CLIP.}
    \label{tab:supp_ablation_layers}
    \vspace{-0.2cm}
\end{table*}

\begin{figure*}[t]
    \centering
    \includegraphics[width=0.8\linewidth]{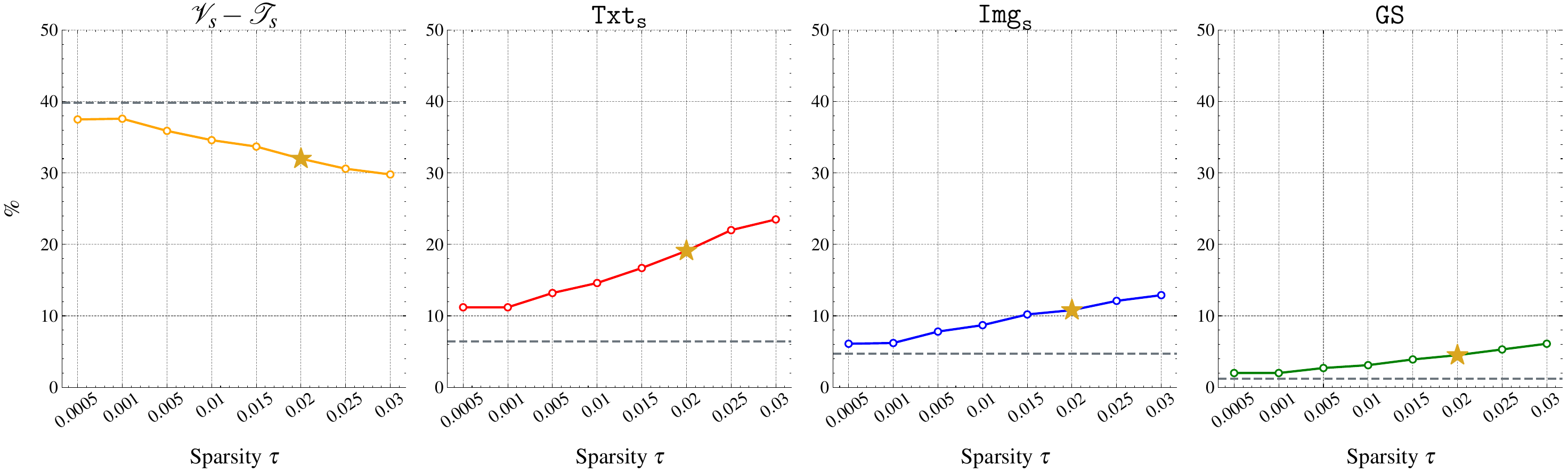}
    \caption{We ablate the sparsity $\tau$ and show the trend of four key metrics \colorbox{orange!20}{$\mathcal{V}_s\text{-}\mathcal{T}_s$}, \colorbox{red!20}{$\mathtt{Txt}_\mathtt{s}$}, \colorbox{blue!20}{$\mathtt{Img}_\mathtt{s}$} and \colorbox{green!20}{$\mathtt{GS}$}. We also provide the original CLIP performance in \colorbox{gray!20}{\protect\raisebox{0.1ex}{- - -}} and the final chosen configuration in \textcolor{Goldenrod}{\Large $\star$}.}
    \label{fig:tau_ablation}
\end{figure*}

\begin{figure*}[t]
    \centering
    \includegraphics[width=\linewidth]{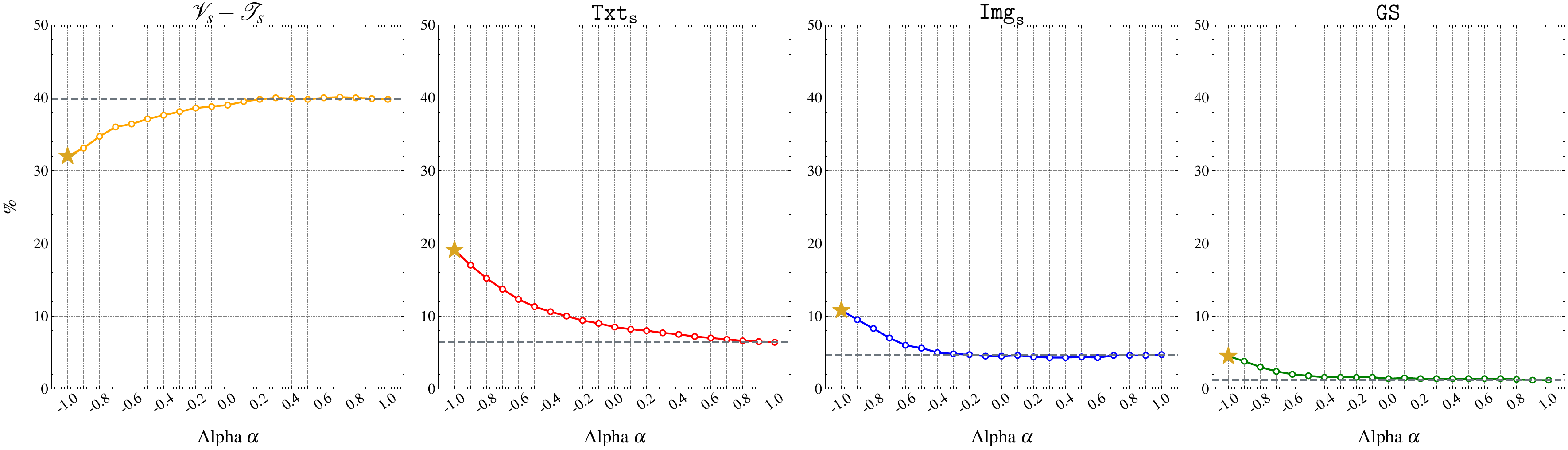}
    \caption{We ablate $\alpha$ and show the trend of four key metrics \colorbox{orange!20}{$\mathcal{V}_s\text{-}\mathcal{T}_s$}, \colorbox{red!20}{$\mathtt{Txt}_\mathtt{s}$}, \colorbox{blue!20}{$\mathtt{Img}_\mathtt{s}$} and \colorbox{green!20}{$\mathtt{GS}$}. We also provide the original CLIP performance in \colorbox{gray!20}{\protect\raisebox{0.1ex}{- - -}} and the final chosen configuration in \textcolor{Goldenrod}{\Large $\star$}.}
    \label{fig:alpha_ablation}
    \vspace{-0.2cm}
\end{figure*}

\section{Ablation \& further details}\label{supp_sec:ablation}
This section extends the ablations by applying \ours i) with different scoring functions, ii) across different layers, iii) at different sparsity levels $\tau$, iv) with different weight scaling factors $\alpha$, and v) applying it individually to each modality. Finally, we provide a societal impact statement and conclude by discussing the baselines.

\subsection{Ablation}

\para{Scoring Function} In Sec. 5.3 of the main paper, we ablate the choice of the scoring function for computing the importance scores $\Phi$. We expand this ablation in~\cref{tab:supp_ablation}, where we add the adaptive selection strategy of Eq. (11) to $\Phi^{\mathtt{uns}}$ and a different scoring function by subtracting the safe scores $\Phi^{\mathtt{sf}}$ to the unsafe ones $\Phi^{\mathtt{uns}}$ (\ie, $\Phi^{\mathtt{uns}}-\Phi^{\mathtt{sf}}$). 

In the first case, adaptively selecting weights using only the unsafe scores $\Phi^{\mathtt{uns}}$ (third \textit{vs} second row) improves the zero-shot performance ($\mathcal{V}_s\text{-}\mathcal{T}_s$ from $24.3\%$ to $30.6\%$), however, the overall safety scores decrease ($\mathtt{GS}$ from $4.6\%$ to $3.4\%$). Moreover, this method makes the model more unsafe when prompted with safe data ($\mathtt{PS}$, from the original $67.5\%$ to $64.3\%$). This behavior is consistent with the same method without adaptivity (second row, dropping from $67.5\%$ to $60.9\%$), showing that including the safe scores $\Phi^{\mathtt{sf}}$ in the scoring function is crucial to disentangle weights associated to unsafe concepts.
For this reason, we explore an alternative score, replacing the merging function $\frac{\Phi^{\mathtt{uns}}}{\Phi^{\mathtt{sf}}}$ with the difference $\Phi^{\mathtt{uns}}-\Phi^{\mathtt{sf}}$. This change aims to reduce the unsafe scores $\Phi^{\mathtt{uns}}$ when high safe scores $\Phi^{\mathtt{sf}}$ are present, allowing for negative values. This strategy (fourth row) shows good zero-shot performance ($31\%$ in $\mathcal{V}_s\text{-}\mathcal{T}_s$), while being capable of improving safety (\eg, $+1.4\%$ in $\mathtt{GS}$). However, this method (i) degrades safety with safe inputs (\ie, $\mathtt{PS}$), and (ii) performs worse than \ours (\eg, $+1.4\%$ in $\mathtt{GS}$ \textit{vs} $+3.3\%$ of \colorbox{ModelLightBlue}{\ours}). Fianlly, the adaptive selection method (fifth row) is capable of improving zero-shot performance preservation ($38.7\%$ in $\mathcal{V}_s\text{-}\mathcal{T}_s$) in line with results of the main paper, however, the scoring function fails in improving safety, with decreases in $\mathtt{Img_s}$ and $\mathtt{GS}$.

\para{Magnitude priors} Following existing works~\cite{sun2023simple}, we treat the weights' magnitudes as priors and test our method by incorporating them into our scoring function, \ie we multiply the final score $\Phi$ by $\left | W\right |$. We ablate this in \cref{tab:prior_ablation} where we apply the priors to each modality independently. This experiment shows that our method is capable of improving safety while preserving zero-shot performance when no prior is applied (first row, \eg $+4.4\%$ $\mathtt{Txt}_\mathtt{s}$). When the prior is applied to the text encoder (second row), \ours greatly improves safety without severe performance degradation (\eg, $+12.7\%$ $\mathtt{Txt}_\mathtt{s}$ and $+6.1\%$ $\mathtt{Img}_\mathtt{s}$). However, when the same procedure is applied to the vision encoder (last two rows), we observe great performance degradation. This shows that the image encoder weights do not serve as a reliable prior for identifying crucial weights.
Thus, during our experiments, we use the magnitude prior only on the text encoder.

\para{Layers} We apply \ours across different layers of both encoders and report the results in~\cref{tab:supp_ablation_layers}. We consider two layers of the self-attention mechanism: the value (Value) and output (Out) projections, and the subsequent MLP: the first (Fc1) and the second (Fc2) fully connected layers. %

When applying the method to the MLP of the text encoder (first two blocks), both Fc1 and Fc2 significantly degrade the zero-shot performance ($\approx 0.0\%$ in $\mathcal{V}_s\text{-}\mathcal{T}_s$) showing that these layers are highly sensitive to pruning. A similar behavior can be observed when pruning the Fc1 of the image encoder (first row of the last two blocks). 

\ours finds unsafe weights while preserving zero-shot performance when applied to the text encoder's value and output projection layers (last two blocks). In the case of the value projection layer (third block), the method improves safety when applied in conjunction with Fc2, value, or output layers of the image encoder (\eg Fc2 $+1.5\%$ in $\mathtt{GS}$). However, it greatly decreases the zero-shot performance ($\mathcal{V}_s\text{-}\mathcal{T}_s$), showing that the value layer of the text encoder is sensitive to pruning and important to the text encoder. 

The best results are achieved when \ours is applied to the output projection layer of the text encoder (last block). We can observe better zero-shot performance preservation while improving safety. Specifically, the best configuration is achieved when pruning simultaneously with the Fc2 of the image encoder, where we improve model safety (\eg $+3.3\%$ in $\mathtt{GS}$) while achieving high performance preservation. A similar performance is obtained by applying \ours to the output projection layer of the image encoder (last row). In this case, we have similar zero-shot performance preservation of the previous configuration, however, we achieve lower safety improvements (\eg $-1.5\%$ in $\mathtt{PS}$). Thus, we prune the Fc2 of the image encoder and the output projection layer of the text encoder across all experiments.

\para{Sparsity} We keep $\alpha=-1$ and ablate the sparsity $\tau$ in a range $\left [10^{-4},10^{-2}\right ]$ and show the trend of four key metrics in~\cref{fig:tau_ablation}. We show the zero-shot performance $\mathcal{V}_s\text{-}\mathcal{T}_s$, the text $\mathtt{Txt}_\mathtt{s}$ and image $\mathtt{Img}_\mathtt{s}$ modality metrics and the group score $\mathtt{GS}$. Additionally, we provide the original CLIP performance as \colorbox{gray!20}{\protect\raisebox{0.1ex}{- - -}} and the final chosen configuration in \textcolor{Goldenrod}{\Large $\star$}.

We observe a consistent and positive increase of $\mathtt{Txt}_\mathtt{s}$, $\mathtt{Img}_\mathtt{s}$, and $\mathtt{GS}$, \ie the more the pruning the better the safety of the model. This consistent trend demonstrates the effectiveness of \ours in discovering weights associated with unsafe concepts. Similarly, we observe the impact of the method on the zero-shot performance of the model in $\mathcal{V}_s\text{-}\mathcal{T}_s$. We recall that a decrease in zero-shot performance is expected as we are pruning meaningful weights. Nevertheless, the decrease is controlled and does not show severe performance degradation, indicating that the model is robust to different sparsity values. As the final configuration, we choose $\tau=0.02$ (\textcolor{Goldenrod}{\Large $\star$}) as it is the best trade-off between zero-shot performance (\eg, $\mathcal{V}_s\text{-}\mathcal{T}_s$) and safety.

\para{Ablation on $\alpha$} We keep $\tau=0.02$, \ie our final configuration for $\tau$, and ablate $\alpha$ in the range from $-1$ to $1$ with intervals of $0.1$. We show the results in~\cref{fig:alpha_ablation} and report the zero-shot performance $\mathcal{V}_s\text{-}\mathcal{T}_s$, the text $\mathtt{Txt}_\mathtt{s}$ and image $\mathtt{Img}_\mathtt{s}$ modality metrics and the group score $\mathtt{GS}$. 

By increasing $\alpha$ towards the original model behavior (from $\alpha=-1$ to $\alpha=1$), we observe that the model's zero-shot performance improves, yet its safety declines, gradually reverting to the unsafe behaviors of the original model. Furthermore, we notice that the image modality $\mathtt{Img}_\mathtt{s}$ metric converges more rapidly to the original behavior compared to the text modality $\mathtt{Text}_\mathtt{s}$, showing the different sensitivities to pruning of the two encoders.
It is important to note that these fine grained details would have been much harder to uncover by using solely retrieval-based metrics.

\begin{table}[t]
    \centering
    \tiny
    \begin{tabular}{c|cc|c|ccccc}
        \toprule
        \multirow{2}{*}{\textbf{Method}}  & \multicolumn{2}{c|}{\textbf{Modalities}} & \textbf{Zero-Shot ($\uparrow$)} & \multicolumn{5}{c}{\textbf{\bench Metrics ($\uparrow$)}} \\ \cmidrule{2-9}
        & $\mathcal{T}$ & $\mathcal{V}$ & $\mathcal{V}_s$-$\mathcal{T}_s$ & $\mathtt{Txt}_\mathtt{s}$ & $\mathtt{Img}_\mathtt{s}$ & $\mathtt{PS}$ & $\mathtt{PU}$ & $\mathtt{GS}$ \\ \midrule
        \rowcolor{gray!20} CLIP   & - & - & 39.8 & 6.4 & 4.7 & 67.5 & 1.7 & 1.2 \\ \midrule
        \multirow{3}{*}{\ours}  & \checkmark & & 32.5 & 18.2 & 10.0 & 66.3 & 4.8 & 4.0 \\
         & & \checkmark              & \textbf{39.4} & 7.6 & 4.2 & \textbf{68.5} & 1.5 & 1.0 \\
         \rowcolor{ModelLightBlue} & \checkmark & \checkmark   & 32.0 & \textbf{19.1} & \textbf{10.9} & 67.8 & \textbf{5.5} & \textbf{4.5} \\
        \bottomrule
    \end{tabular}
    \caption{We ablate \ours across modalities.}
    \label{tab:modalities_ablation}
\end{table}

\para{Modality Ablation} We further ablate \ours by applying it individually to each modality in~\cref{tab:modalities_ablation}. This experiment aims to analyze the relative contributions of each modality to the overall model performance. We observe that \ours significantly improves the text encoder nearly achieving its best results. When applied to the image encoder, the improvements are lower with $+1.2\%$ in $\mathtt{Txt}_\mathtt{s}$ and $+1.0\%$ in $\mathtt{PS}$ and a slight decrease in $\mathtt{Img}_\mathtt{s}$ and $\mathtt{PU}$. This further shows the complexity of pruning meaningfully the image encoder. Nevertheless, when \ours is applied to both encoders, we obtain the best performance, showing the effectiveness of pruning both the image and text encoders simultaneously. 

\subsection{Societal Impact}
\ours enhances safety across different VLM architectures (Tab. 4) and LLaVA~\cite{liu2023llava} (Tab. 5). By improving safety, \ours helps mitigate harmful behaviors that may be transferred to downstream models (\eg, LLaVA), thus may help to reduce the spread of harmful content. As AI systems become more widely used, ensuring their safety is crucial for advancing responsible and ethical AI development. Nevertherless, it is important to consider potential negative aspects. In this study, we focused on improving safety by localizing and manipulating unsafe weights, demonstrating safety improvements across several settings. However, the same procedure could, in principle, be applied to safe weights with the malicious intent of making the model more unsafe. While we strongly oppose such behavior, as it contradicts our ethical standards, we note that \ours, in its current form, is specifically designed to localize unsafe weights and cannot be directly used for such a malicious objective. Achieving the opposite goal would require significant modifications to our method (\eg, rethinking the scoring function). Ultimately, \ours contributes to safer AI systems, advancing the ethical use of machine learning models while mitigating the risk of harmful behaviors.

\subsection{Baselines}
Aside from \ours, this work introduces two gradient-based pruning baselines \textit{\baselineOne} and \textit{\baselineTwo}. The workflow is consistent across both baselines: after computing the objective function, the weights with the highest gradient magnitude are pruned~\cite{NEURIPS2023_44956951, xu2024besa, das2023beyond}. Both baselines include a sparsity parameter (\ie, the layer-wise percentage of weights to prune), which is tuned through experiments on the held-out validation set, as detailed in Sec. 5.1 of the main paper. The core difference between the two baselines is the objective function used for weight localization. Given a sample $(v_s, v_u, t_s, t_u)\in\mathcal{D}$, \textit{\baselineOne} leverages a loss function to align safe content (\ie, $v_s$, $t_s$) with unsafe one (\ie, $v_u$, $t_u$) while simultaneously distancing safe content of the opposite modality (\eg, $v_s$ \textit{vs} $t_s$). Let us denote with $f_\mathtt{Txt}$ the text encoder receiving as input text in the space $\mathcal{T}$, \ie $f_\mathtt{Txt}:\mathcal{T}\rightarrow\mathbb{R}^d$, and outputs the textual representation with dimensionality $d$. Similarly, the image encoder encodes an image from the space $\mathcal{V}$ and outputs its representation, \ie, $f_\mathtt{Img}:\mathcal{V}\rightarrow\mathbb{R}^d$. Since our goal is to receive gradients only for redirecting safe samples (\eg, $t_s^i$), we freeze the encoders used to process the unsafe ones (\eg, $v_u^i$) and the safe sample of the opposite modality (\eg, $v_s^i$). Specifically, the objective function for the text encoder is defined as:
\begin{align}
    \mathcal{L}_\mathtt{Txt}=\frac{1}{|\mathcal{D}|}\sum_{i=1}^{|\mathcal{D}|} \bigg( & D_c(f_\mathtt{Txt}(t_s^i), f_\mathtt{Txt}^\mathtt{Base}(t_u^i)) \\
    & - D_c(f_\mathtt{Txt}(t_s^i), f_\mathtt{Img}^\mathtt{Base}(v_s^i)) \bigg)
\end{align}
where we denote with $D_c$ the cosine distance and the $\mathtt{Base}$ superscript the frozen encoders. 
The same procedure is carried out on the image encoder:
\begin{align}
    \mathcal{L}_\mathtt{Img} = \frac{1}{|\mathcal{D}|} \sum_{i=1}^{|\mathcal{D}|} \bigg( & D_c(f_\mathtt{Img}(v_s^i), f_\mathtt{Img}^\mathtt{Base}(v_u^i)) \\
    & - D_c(f_\mathtt{Img}(v_s^i), f_\mathtt{Txt}^\mathtt{Base}(t_s^i)) \bigg)
\end{align}
\textit{\baselineOne} minimizes the above losses to prune the two encoders separately. For example, when pruning $f_\mathtt{Txt}$, \textit{\baselineOne} uses $\mathcal{L}_\mathtt{Txt}$ and prunes the weights receiving the highest magnitude gradient.

\textit{\baselineTwo} leverages the contrastive redirection loss from~\cite{poppi2024removing}, but with an adaptation to reverse its effect: it pulls safe representations closer to unsafe ones. The key difference between \textit{\baselineOne} and \textit{\baselineTwo} is that \textit{\baselineTwo} uses a contrastive loss based on cross-entropy, while \textit{\baselineOne} employs a simpler loss function that operates between individual samples.

\begin{figure*}
    \centering
    \includegraphics[width=0.89\textwidth]{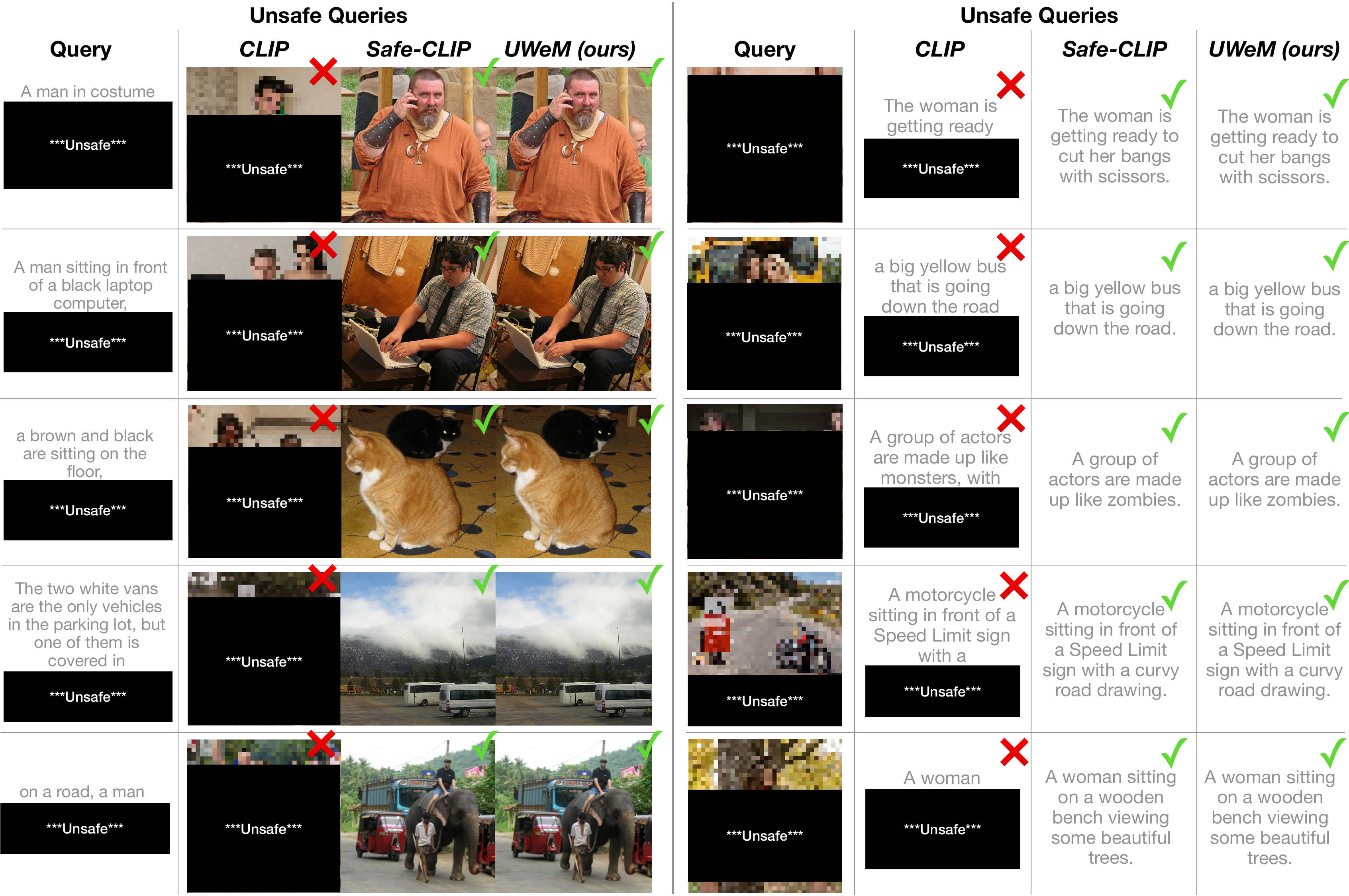}
    \caption{Qualitative results by prompting CLIP~\cite{radford2021learning}, \safeclip~\cite{poppi2024removing}, and \ours with \textit{unsafe} data.}
    \label{fig:qualitatives_unsafe_query}
\end{figure*}

\begin{figure*}
    \centering
    \includegraphics[width=0.89\textwidth]{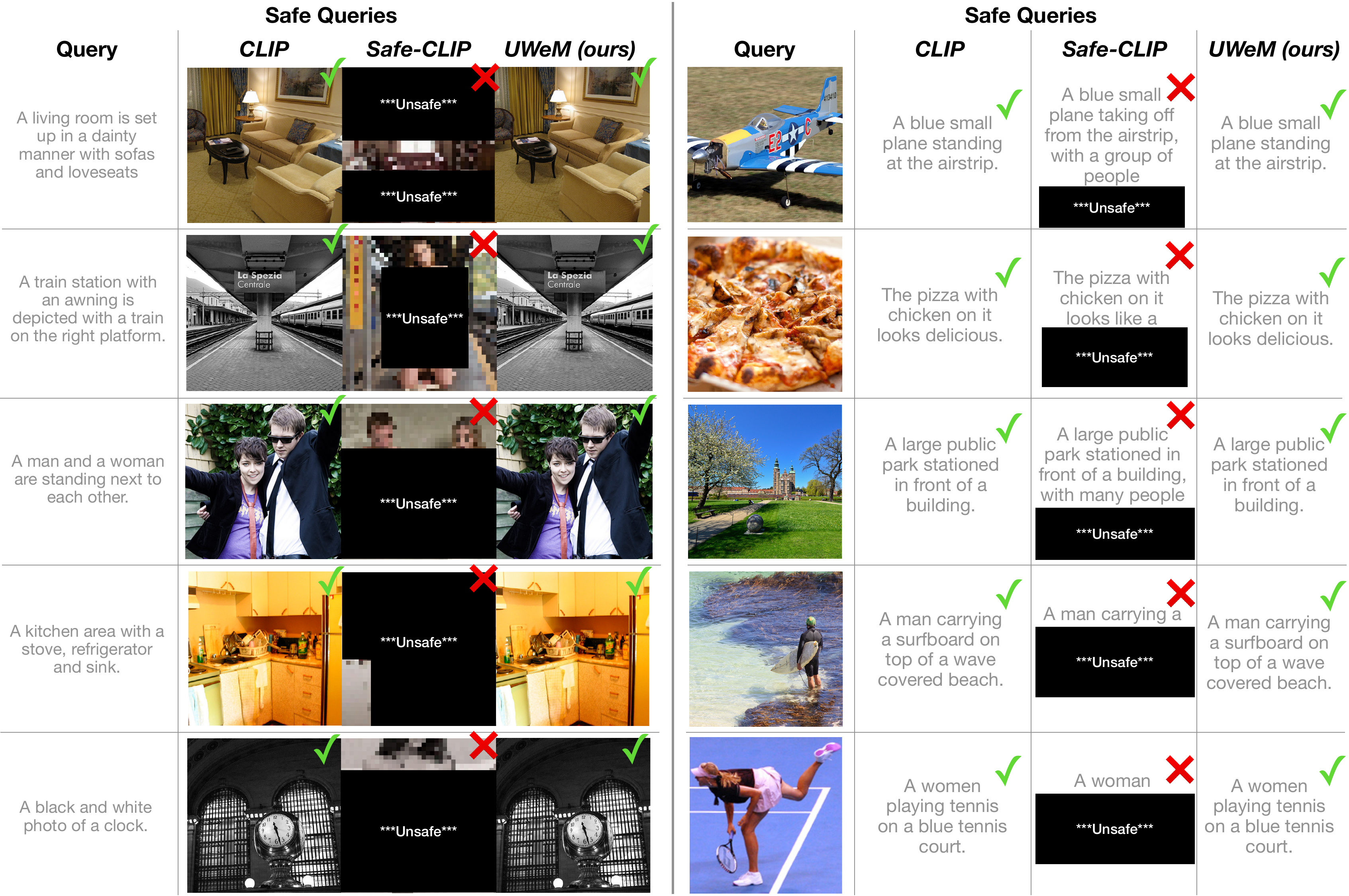}
    \caption{Qualitative results by prompting CLIP~\cite{radford2021learning}, \safeclip~\cite{poppi2024removing}, and \ours with \textit{safe} data.}
    \label{fig:qualitatives_safe_query}
\end{figure*}

\begin{figure*}
    \centering
    \includegraphics[width=0.89\textwidth]{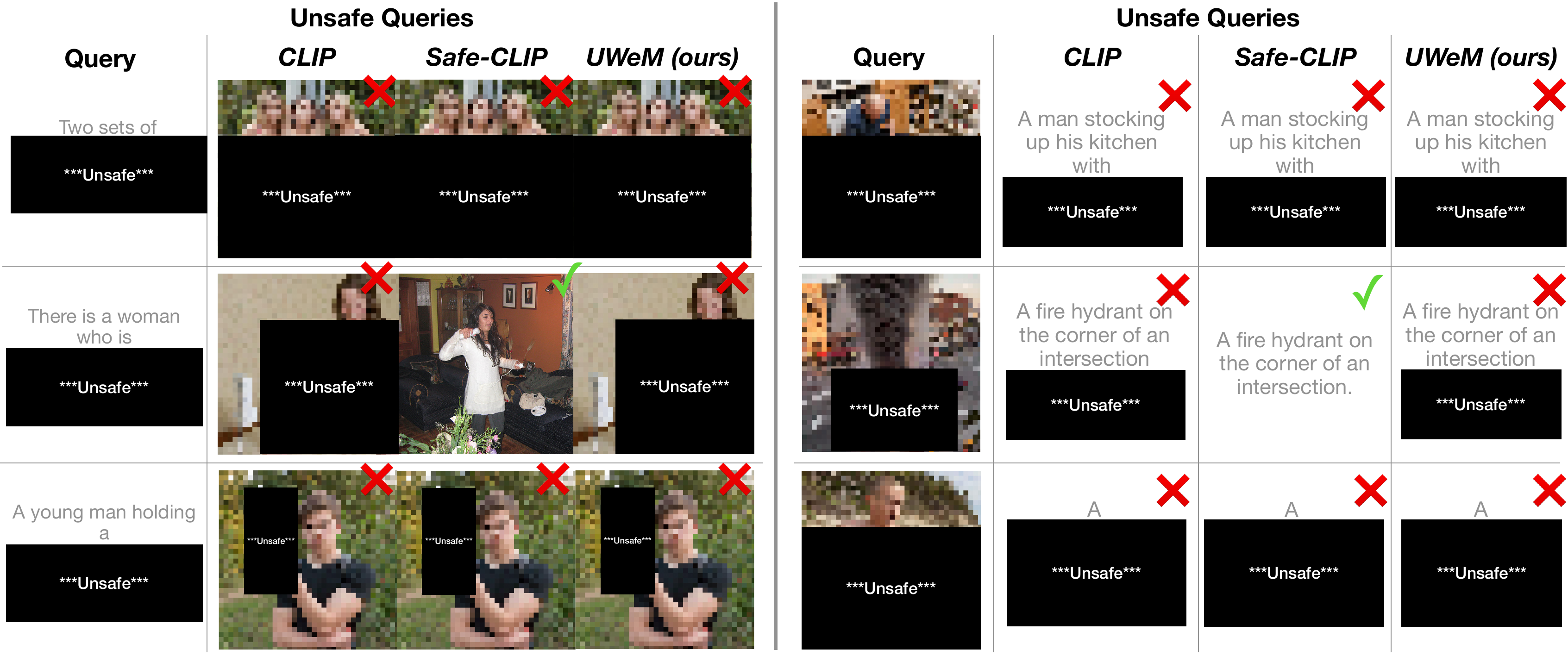}
    \caption{Failure cases of the proposed method \ours.}
    \label{fig:failure_cases}
\end{figure*}

\section{Qualitative Results}\label{supp_sec:qualitatives}
We discuss further qualitative results in this section. We prompt CLIP~\cite{radford2021learning}, \safeclip~\cite{poppi2024removing}, and \ours with \textit{unsafe} and \textit{safe} data and present the results in~\cref{fig:qualitatives_unsafe_query} and~\cref{fig:qualitatives_safe_query}. 

When prompted with unsafe data (\cref{fig:qualitatives_unsafe_query}), CLIP demonstrates unsafe behaviors by retrieving harmful content, highlighting its vulnerability to unsafe prompts and the critical need for effective mitigation strategies. In contrast, both \safeclip and \ours consistently exhibit robust and reliable performance across various prompts in both image and text modalities, showing their effectiveness in mitigating the original unsafe behavior of CLIP. 

However, when prompted with safe queries (\cref{fig:qualitatives_safe_query}), \safeclip fails to prefer safe data and retrieves unsafe content, showing the limitations of training-based techniques. On the other hand, CLIP shows a safe behavior by retrieving safe content and, importantly, \ours preserves this feature, demonstrating its effectiveness in addressing both unsafe and safe query scenarios. 

\para{Failure Cases} In~\cref{fig:failure_cases}, we show failure cases where models are prompted with unsafe data. In these cases, \ours struggles to enhance CLIP’s safety, retaining its original harmful behavior. Similarly, \safeclip, despite being specifically fine-tuned for such scenarios (\ie, unsafe queries), also encounters challenges in mitigating unsafe responses (\eg, first and third row). The limitations of both \ours and \safeclip are closely tied to the inherent unsafety of CLIP, which exhibits a concerning predisposition to favor unsafe content, \eg, with a likelihood of $93.6\%$ for the text modality ($100-\mathtt{Txt}_\mathtt{s}$, first row in \cref{tab:supp_ablation_layers}) and $95.3\%$ for the image modality ($100-\mathtt{Img}_\mathtt{s}$). While both methods demonstrate the ability to reduce this unsafe behavior, removing it entirely is challenging. This further highlights the complexity of mitigating harmful responses and reaffirms that addressing such behaviors remains an open and critical area of research.

\end{document}